\documentclass[11pt,a4paper]{article}
\usepackage[margin=25mm]{geometry}
\usepackage[T1]{fontenc}
\usepackage{lmodern,amsmath,amssymb,graphicx,booktabs,multirow}
\usepackage[table]{xcolor}
\usepackage{caption}
\usepackage[numbers,sort&compress]{natbib}
\usepackage{tcolorbox}
\usepackage{graphicx}
\tcbuselibrary{skins,breakable}
\usepackage{fancyhdr}
\usepackage{fontawesome5}
\definecolor{paperGreen}{HTML}{27854C}
\definecolor{titleGold}{HTML}{94712F}
\definecolor{paperTint}{HTML}{F2F8F4}
\usepackage[colorlinks=true,allcolors=paperGreen]{hyperref}
\usepackage[capitalize]{cleveref}

\renewenvironment{abstract}{\par\vspace{0.7em}\normalsize\color{black}}{\par}
\begin{document}
\begin{tcolorbox}[enhanced,breakable,colback=paperTint,colframe=paperTint,boxrule=0pt,arc=4mm,left=6mm,right=6mm,top=5mm,bottom=5mm]
{\LARGE\bfseries 
\textcolor{titleGold}{Touch\'{e}}World Tech Report\\
\texorpdfstring{%
    \raisebox{-0.15em}{%
      \includegraphics[height=1.2em]{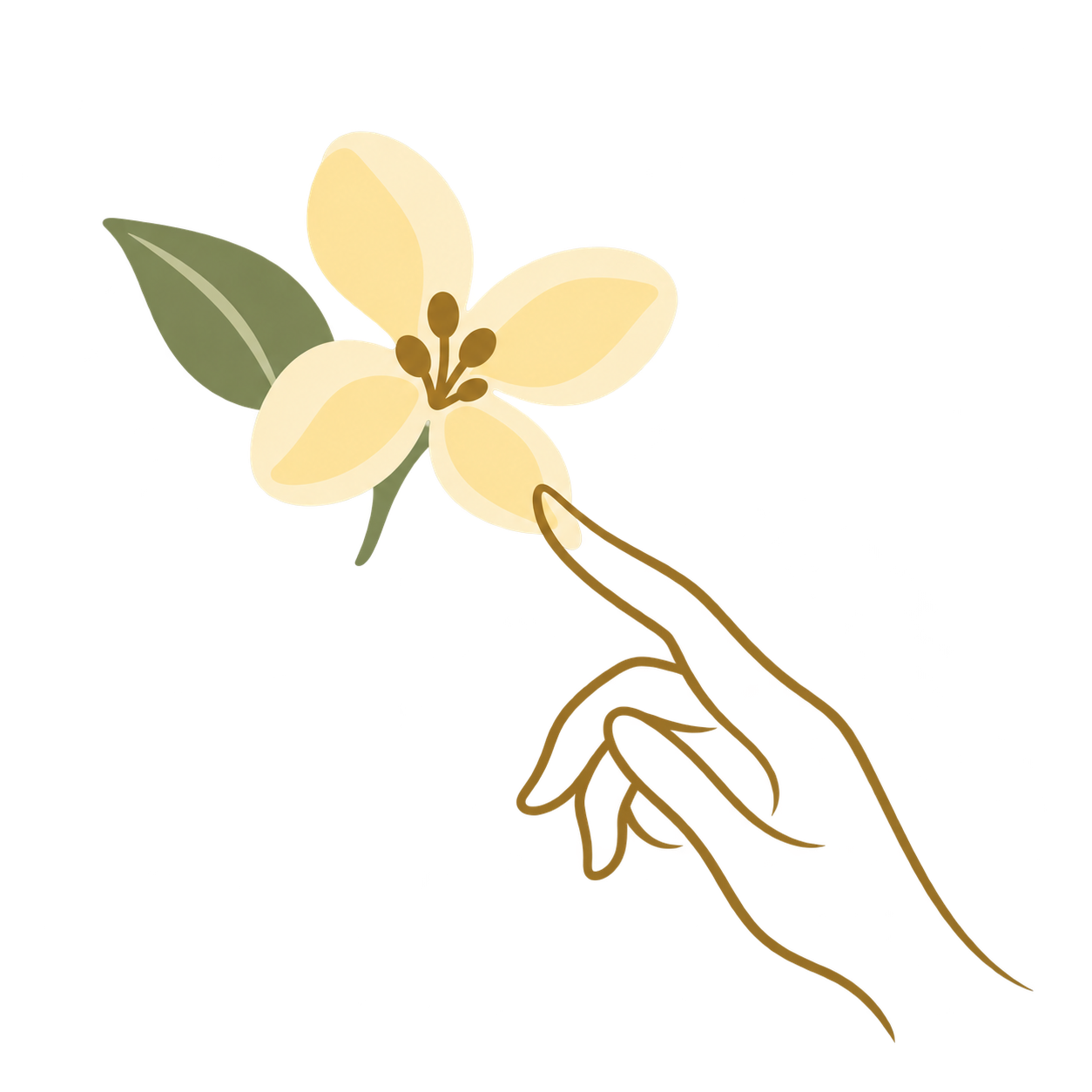}%
    }\hspace{0.3em}%
  }{}%
\textcolor{titleGold}{T}owards \textcolor{titleGold}{O}nline \textcolor{titleGold}{U}nified \textcolor{titleGold}{C}amera--\textcolor{titleGold}{H}and\\
\textcolor{titleGold}{E}xploration in Generative Worlds\par}
\vspace{0.7em}

{\color{paperGreen}\large 
\begin{flushleft}
  {\large
  Jiaao Yu$^{3,*}$ \quad
  Yanjian Zhang$^{1,2,*,\dagger}$ \quad
  Haoyang Jiang$^{4}$ \quad \\
  \vspace{3pt}
  Guillaume Wisniewski$^{2}$ \quad
  Nadi Tomeh$^{1}$ \quad
  Shijie Jia$^{4}$\\
  }
  \vspace{6pt}
  {\small
    $^{1}$Universit\'{e} Sorbonne Paris Nord, LIPN, CNRS \quad
    $^{2}$Universit\'{e} Paris Cit\'{e}, LLF, CNRS \quad \\
    $^{3}$East China Normal University \quad 
    $^{4}$Dalian Jiaotong University
  }
  \vspace{-5pt}

\end{flushleft}
\par}

\begin{abstract}
Recent video world models allow users to explore generated environments through camera navigation. Continuous hand motion provides a fine-grained action channel for interacting with the people and objects encountered during exploration. We present Touch\'{e}World, a framework that integrates gesture-driven interaction into a camera-controlled video world model. Given an initial image and a scene-description prompt, the model generates video conditioned on temporally aligned camera and hand controls. We construct joint camera--hand training data from heterogeneous egocentric videos through annotation completion, consistency-based filtering, and temporal alignment. Building on a pretrained camera-controlled world model, we combine spatial hand-skeleton conditioning with numerical hand-motion features through parameter-efficient adaptation. Causal adaptation and few-step distillation support chunk-level online generation, allowing users to update their controls based on generated visual feedback. We further introduce Touch\'{e}Bench to evaluate hand-motion following, camera--hand coordination, online interaction, and object response. Together, these components establish a framework for studying continuous exploration and hand-driven interaction within generative worlds.
\end{abstract}

\vspace{5pt}
{\color{paperGreen}\large 
  {\footnotesize
    $^{*}$Equal contribution.
    \qquad
    $^{\dagger}$Project lead.
  }
\par}

\vspace{5pt}

\begin{center}
  \small
  \href{https://haoyangjang.github.io/toucheworld/}
       {\color{titleGold}\faGlobe\ Project Page}
  \quad
  \href{https://huggingface.co/YJAYJA/ToucheWorld}
       {\color{titleGold}\faDownload\ Model Weights}
  \quad
  \href{https://huggingface.co/datasets/YJAYJA/ToucheBench}
       {\color{titleGold}\faDatabase\ Dataset}
  \quad
  \href{https://github.com/yjainqdc/ToucheWorld/}
       {\color{titleGold}\faGithub\ Code}
\end{center}

\end{tcolorbox}
\begin{figure}[htbp]
\centering
\includegraphics[width=\linewidth]{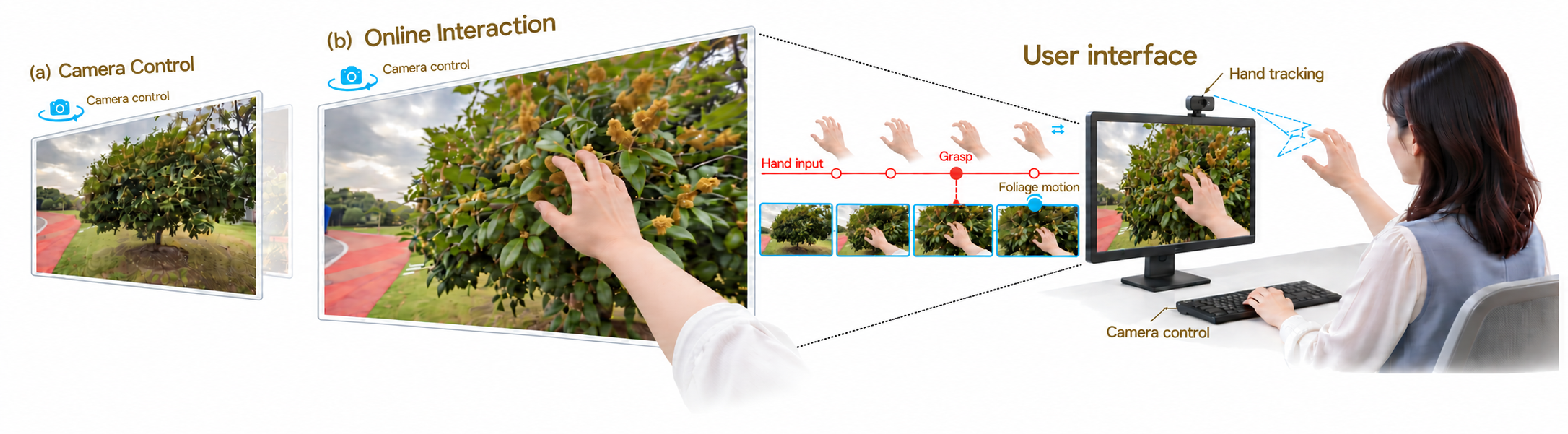}
\caption{
From camera-guided exploration to continuous hand--world interaction.
Camera control allows users to navigate generated environments, while hand motion provides a fine-grained action channel for interacting with encountered scene contents.
In the illustrated example, the user approaches an osmanthus tree and reaches into its canopy, producing hand contact and corresponding changes in the foliage.
With scene-only text conditioning, camera and hand inputs guide the unfolding interaction, while causal chunkwise generation incorporates updated controls into subsequent visual output.
}
\label{fig:onecol}
\end{figure}
\section{Introduction}
\label{sec:intro}

\begin{figure*}[t]
  \centering
\includegraphics[width=1.0\linewidth]{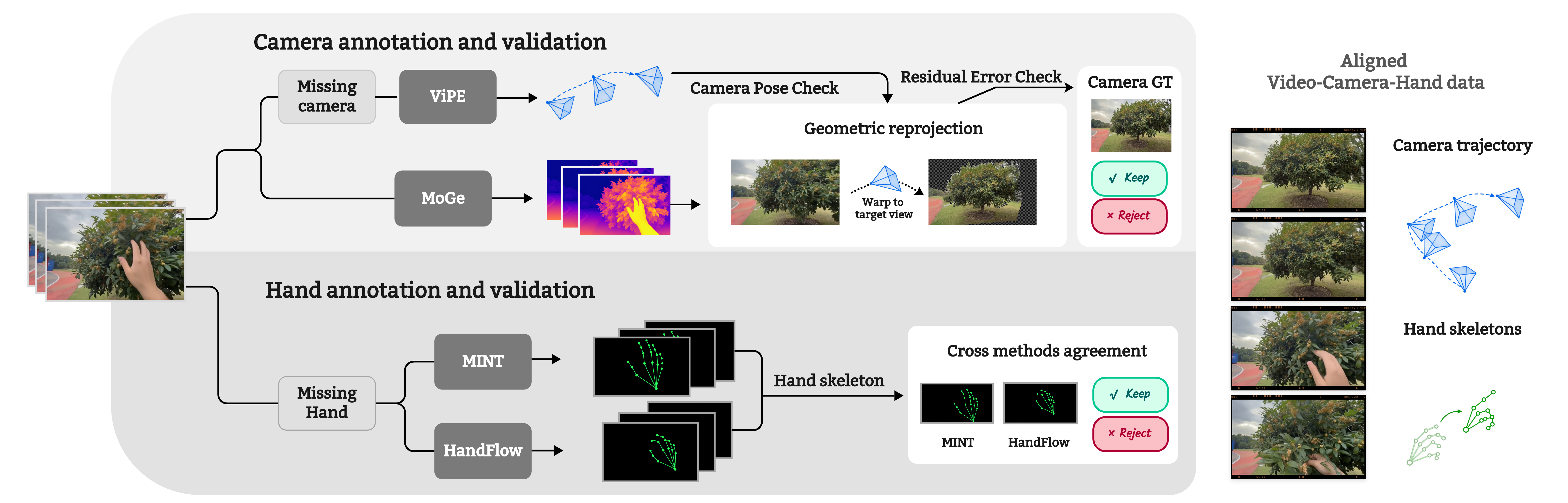}
\caption{Joint camera--hand data construction through annotation completion, consistency-based filtering, and temporal alignment.}
   \label{fig:data}
\end{figure*}

Video world models are turning generated visual content into environments that users can explore. Advances in camera-conditioned generation \cite{zhang2026unified,li2026cameras,he2026matrixgame20opensourcerealtime} and streaming inference \cite{zhu2026causal,zhao2026causalforcingscalablefewstep} allow users to move through scenes, choose their viewpoints, and explore worlds as they unfold~\cite{zhang2026echowm,zhu2026causal,zhao2026causalforcingscalablefewstep,huang2026self}. Exploration, however, naturally invites participation. Users may wish not only to observe what they encounter, but also to act on it. Consider walking through a generated courtyard, approaching a sweet-scented osmanthus tree, reaching out to touch its branches, and then continuing along the path. Supporting this process requires camera-guided navigation and hand-driven interaction to work together, with the generated world responding to user actions as exploration continues.

Continuous hand motion provides a natural interface for such participation. Camera control determines how users observe the world, while hand position, orientation, and articulation provide a fine-grained action channel for influencing its contents. Users can express actions such as reaching, touching, pushing, and withdrawing, and adjust their subsequent behavior based on generated visual feedback. We organize this process as online interaction: the system receives user controls while producing visual output and incorporates updated controls into subsequent video chunks, allowing users to decide their next action through ongoing observation.

Recent work provides important foundations through hand--object dynamics generation, joint hand and camera control, and streaming interaction~\cite{kim2026dexterous,xie2026generatedrealityhumancentricworld,wang2026hand2worldautoregressiveegocentricinteraction}. The supported range of scenes and interactions nevertheless remains limited, particularly when manipulation is embedded within continuous navigation and exploration. We focus on the continuity across these activities: a user navigates toward an object, reaches out to manipulate it, withdraws their hand, and resumes exploration. Across these transitions, viewpoint control, hand motion, and scene responses must remain coordinated, while subsequent generation carries forward the state established by earlier actions. This calls for jointly considering the data, generation process, and evaluation of continuous camera--hand interaction.

Three challenges arise. First, joint control requires reliable supervision that separates viewpoint changes from action-induced scene dynamics. Camera motion, hand articulation, and their visual consequences must be accurately annotated and temporally aligned, despite incomplete or inconsistent estimates in egocentric videos. Second, online generation requires causal processing of both controls and visual history: each generated chunk must respond to available camera and hand inputs without relying on future actions. Third, interaction quality extends beyond visual similarity. A generated hand may follow its controls while the object remains unresponsive, and camera motion may produce apparent hand--object alignment without the intended manipulation. Evaluation must therefore distinguish hand following, joint control, and action-dependent scene responses.

To this end, we present a framework that integrates continuous hand interaction into a camera-controlled video world model. We first construct joint camera--hand training data from heterogeneous egocentric videos, completing missing annotations, filtering camera and hand estimates through consistency checks. Building on a pretrained camera-controlled model, we combine spatial hand-skeleton conditioning with numerical motion features through parameter-efficient adaptation. We further impose consistent causal constraints on hand conditioning and visual generation, and apply few-step distillation to accelerate chunk-wise synthesis. The resulting model incorporates updated camera and hand controls into subsequent chunks, connecting exploration and manipulation within a shared generated history.

We further introduce Touch\'{e}Bench to evaluate four aspects of interactive world generation: hand-motion following, camera--hand coordination, online interaction, and object response. Reference-based evaluation on HOI4D~\cite{Liu_2022_CVPR} measures hand-control accuracy and object-response fidelity, while purpose-recorded scenarios assess joint camera--hand control and action-dependent continuations from shared contexts. Alongside Touch\'{e}Bench, we evaluate video quality on unseen scenes from EgoStandard~\cite{lightwheel2026egosuite} and EgoLife~\cite{Yang_2025_CVPR}, and on the out-of-domain HOT3D dataset~\cite{Banerjee_2025_CVPR}. Comparisons with external methods and across our bidirectional teacher, causal model, and four-step distilled model characterize visual quality and interaction performance across generation stages. Ablation studies further examine the effects of hand-weighted training on hand following and interaction outcomes.

Our contributions are threefold:

\begin{itemize}
    \item \textbf{Joint camera--hand training data.}
    We construct a data pipeline that completes missing camera and hand annotations, filters estimates through consistency checks, and aligns video and control signals to provide joint supervision for camera-guided exploration and hand-driven interaction.

    \item \textbf{A framework for online camera--hand interaction.}
    We extend a camera-controlled video world model with spatial skeleton and numerical motion conditioning. Parameter-efficient adaptation, causal training, and few-step distillation enable chunk-level online generation with continuously updated camera and hand controls.

    \item \textbf{Touch\'{e}Bench and experimental evaluation.}
    We introduce a benchmark covering hand-motion following, camera--hand coordination, online interaction, and object response. Comparisons and ablation studies evaluate these capabilities alongside video quality and examine the effects of causal adaptation, few-step distillation, and hand-weighted training.
\end{itemize}

\begin{figure*}[t]
  \centering
\includegraphics[width=1.0\linewidth]{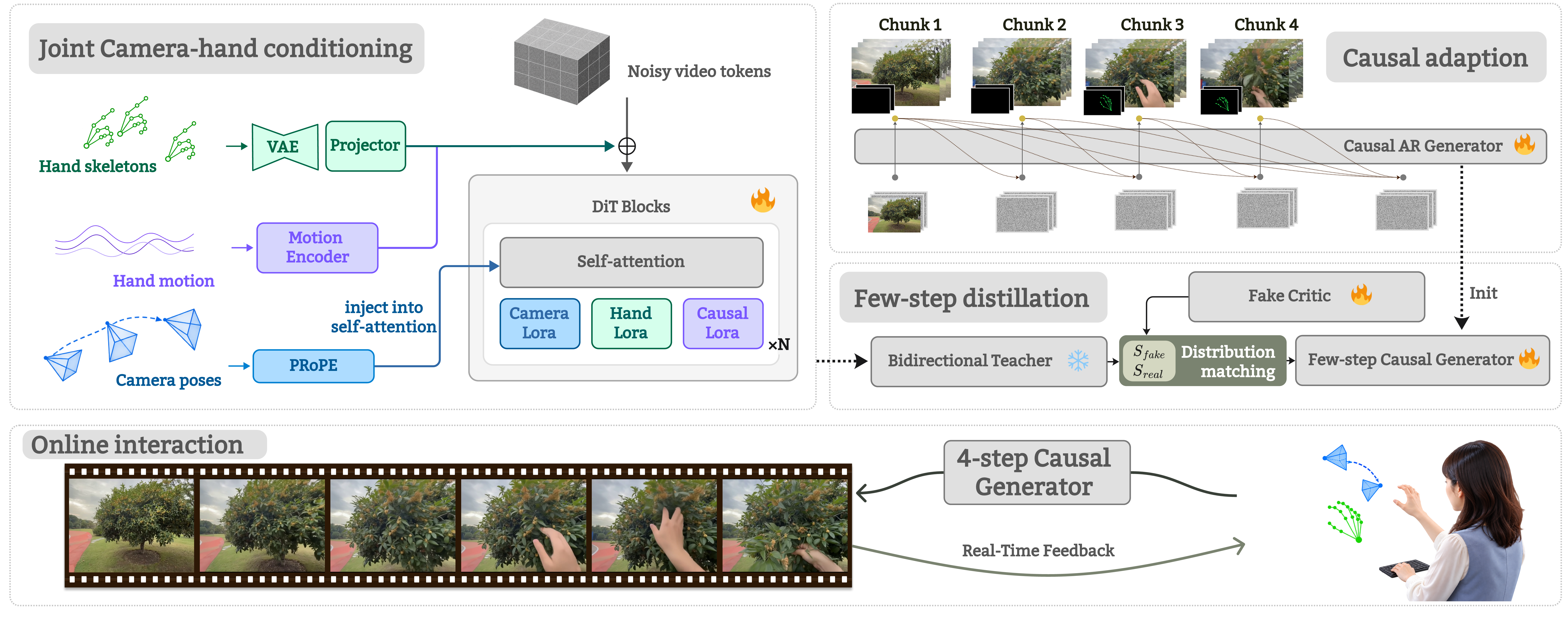}
    \caption{Overview of Touch\'{e}World: joint camera--hand conditioning, causal adaptation, few-step distillation, and online interaction with updated controls.}
   \label{fig:main}
\end{figure*}

\section{Related work}
\label{sec:formatting}

\subsection{Action-Conditioned Video World Models}
Action-conditioned video world models enable generated environments to respond to user control and evolve continuously by incorporating action signals into the video generation process. 
While early works primarily utilized discrete keystrokes or latent actions as control signals \cite{rui2026biwmadvancingopensourceinteractive,zhang2025matrixgameinteractiveworldfoundation,he2026matrixgame20opensourcerealtime}, recent research has introduced continuous camera trajectories to support more flexible navigation and viewpoint changes \cite{team2026dreamx,huang2026solarwmopendatascalable,zhang2026echowm,zhao2026minwm,gao2026infiniteworldsversatileinteractions,che2025gamegen,team2026alayaworld,sun2025worldplay}. 
To link camera motion with spatial changes in the generated imagery, existing methods have explored various geometric representations and conditioning mechanisms. For instance, LingBot-World 2.0~\cite{gao2026infiniteworldsversatileinteractions} employs Plücker embeddings to encode pixel-level observation rays and injects camera conditions via adaptive layer normalization; SolarWM~\cite{huang2026solarwmopendatascalable}  utilizes fused-PRoPE~\cite{li2026cameras} to integrate projection geometry transformations, derived from camera intrinsics and extrinsics into self-attention computations; and EchoWM~\cite{zhang2026echowm} adopts relative UCPE~\cite{zhang2026unified} to encode geometric relationships between local ray coordinate systems, thereby supporting continuous trajectory control. 

Beyond viewpoint navigation, recent work explores control \cite{bruce2024genie,gu2026worldcraft,guo2025mineworld,ICLR2025_b71ecea2} interfaces for character behavior, scene events, and object dynamics. GameGen-X \cite{che2025gamegen} and Hunyuan-GameCraft-2 \cite{tang2025hunyuan} combine instruction-based control with game interaction, enabling users to specify actions and changes to scene contents. LongLive enables interactive streaming video generation and introduces KV-recache to refresh cached states when text prompts change~\cite{yang2025longlive}. H3-World \cite{chen2026h3} explores language-based action control on MiniMax-H3 \cite{minimax2026h3} by converting character and camera actions into temporally aligned textual instructions~\cite{chen2026h3}. Complementary approaches introduce spatial control signals: MotionStream \cite{shin2026motionstream} supports interactive motion guidance, while Force Prompting \cite{gillman2026force} and StreamForce \cite{wang2026streaming} condition generation on physical forces, with StreamForce extending force control to causal streaming generation. Code World Model explores an alternative formulation in which a coding agent maintains executable world states and compiles them into proxy videos that condition visual generation~\cite{chen2026codeworldmodelcoding}. These studies broaden the ways users can influence generated worlds. Our work investigates continuous hand motion as an action interface, jointly with camera control, to connect exploration and manipulation within an ongoing generation process.

\subsection{Hand-Conditioned Video Generation and Interaction Evaluation}

Hand-conditioned video generation models how hand motion influences scene contents. InterDyn~\cite{akkerman2025interdyn} conditions video generation on driving-motion cues, including hand-mask sequences, to synthesize the resulting interaction dynamics. Dexterous World Models \cite{kim2026dexterous} combines hand-mesh renderings with static-scene videos rendered along a specified camera trajectory to model action-induced scene changes. Recent egocentric approaches further investigate joint hand and viewpoint control: Generated Reality \cite{xie2026generatedrealityhumancentricworld} conditions generation on tracked head and hand poses and supports autoregressive interaction. HandsOnWorld \cite{chen2026handsonworld} learns from unconstrained monocular videos and introduces camera-disentangled hand control.
Hand2World \cite{wang2026hand2worldautoregressiveegocentricinteraction} and JCV \cite{zhang2026controllable} primarily study hand--object interaction in tabletop manipulation settings. While these settings capture local interactions, they provide limited coverage of the transitions between navigation and manipulation.
Related interaction-generation methods also investigate structural supervision: SCAR jointly generates videos and structure- and contact-aware representations to improve hand--object interaction synthesis~\cite{yan2025openworldhandobjectinteractionvideo}. Building on these developments, we focus on connecting camera-guided exploration and hand-driven manipulation within a continuous generation process, supported by joint camera--hand data construction and causal control.

Interaction evaluation examines both control fidelity and the consequences of actions. Existing methods assess visual quality, hand-motion agreement, and generated interaction behavior under their respective conditioning settings~\cite{akkerman2025interdyn,kim2026dexterous,xie2026generatedrealityhumancentricworld,wang2026hand2worldautoregressiveegocentricinteraction}. Complementing control-conditioned evaluation, Ego2Act evaluates goal-directed egocentric manipulation from an initial image and a high-level goal, assessing task completion and physical plausibility~\cite{irawan2026ego2actevaluatinggoaldirectedmanipulation}.
Continuous camera--hand interaction additionally calls for evaluating how newly introduced controls affect subsequent generation and how hand actions operate alongside navigation. Touch\'{e}Bench brings these aspects together through hand-motion following, camera--hand coordination, online intervention response, and object response. Paired trajectory evaluation and action interventions from shared contexts relate control accuracy to the resulting interaction outcomes.
\section{Method}
\label{sec:method}

\begin{figure*}[t]
  \centering
\includegraphics[width=1.0\linewidth]{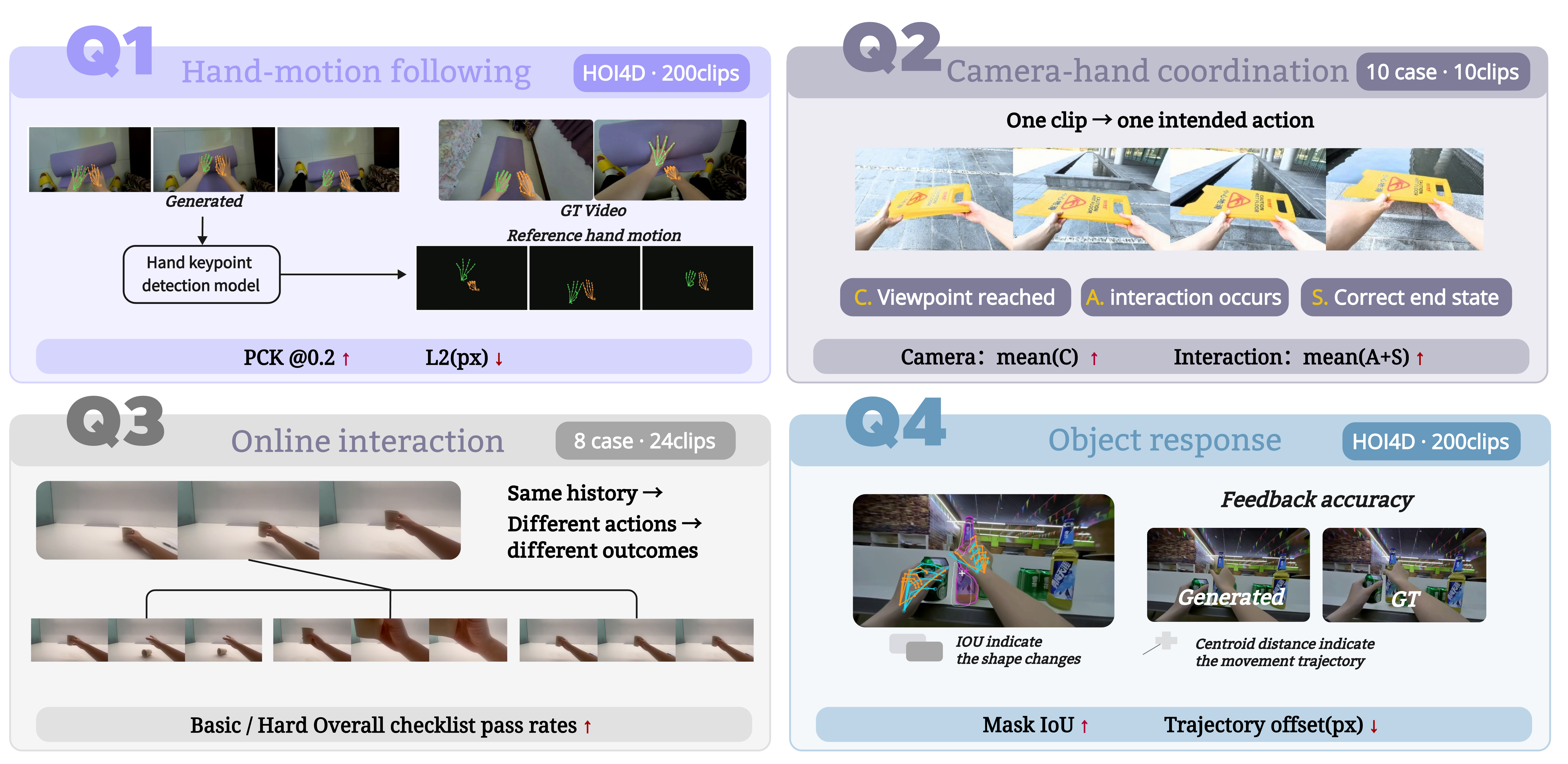}
\caption{Touch\'{e}Bench evaluates hand-motion following (Q1), camera--hand coordination (Q2), online interaction (Q3), and object response (Q4).}
   \label{fig:bench}
\end{figure*}

We present a video world modeling framework that integrates continuous hand interaction with camera-guided exploration. Building on a camera-controlled image-to-video model, we introduce complementary skeleton and motion conditions, followed by causal adaptation and few-step distillation. Figure~\ref{fig:data} illustrates joint data construction, while Figure~\ref{fig:main} summarizes conditioning, training, and online interaction.

\subsection{Problem Formulation}
\label{sec:formulation}

Given an initial image $I_0$ and a scene-description prompt $p$, we generate a continuous video conditioned on camera controls $\mathcal{C}$ and hand controls $\mathcal{H}$. Camera controls specify the viewing trajectory, while hand controls specify hand poses and motion at the corresponding video time positions. Together, they allow users to navigate the generated environment and interact with its contents.

We generate video chunks $X_1,\ldots,X_K$ sequentially:
\begin{equation}
\begin{aligned}
&p_\theta(X_{1:K}\mid I_0,p,\mathcal{C},\mathcal{H})\\
&=\prod_{k=1}^{K}p_\theta\!\left(
X_k\mid I_0,p,X_{<k},
\mathcal{C}_{\leq k},\mathcal{H}_{\leq k}
\right),
\end{aligned}
\label{eq:causal_factorization}
\end{equation}
where $\mathcal{C}_{\leq k}$ and $\mathcal{H}_{\leq k}$ denote the controls available when generating chunk $k$. Within each chunk, hand-control latents are temporally aligned with the corresponding video latent positions.

The system receives user controls while producing visual output and incorporates updated controls into subsequent chunks. This chunk-level online interaction allows users to adjust camera movement and hand actions based on generated visual feedback, connecting exploration and interaction within a shared video history.

\subsection{Data Construction}
\label{sec:data_pipeline}

As shown in Figure~\ref{fig:data}, we construct aligned video--camera--hand data from heterogeneous egocentric videos. We retain valid existing annotations and complete missing camera and hand annotations through separate estimation and validation branches.

For videos without camera annotations, we estimate camera poses using ViPE~\cite{huang2025vipevideoposeengine} and depth using MoGe2~\cite{NEURIPS2025_336572db}. We reproject source frames into target views using the estimated geometry and
compare them with the observed target frames. Reprojection consistency and trajectory-quality checks are used to filter unreliable camera estimates.

For videos without hand annotations, we independently estimate hand poses using MINT~\cite{zhu2026mintunifiedmodelworldspace} and HandFlow~\cite{xu2026handflowfullygenerative4d} and reject samples with inconsistent predictions. The retained annotations are converted into skeleton renderings and numerical hand features for joint conditioning.

We temporally align video frames, camera poses, and hand annotations, updating camera intrinsics and image-space hand coordinates consistently with resizing and cropping.
Camera trajectories are expressed relative to the initial frame. Alongside interaction clips, we include clips with hands entering an initially hand-free view and static clips with and without hands, covering both interaction and periods without hand motion.

We plan to release the constructed dataset to support further research on joint camera--hand control in video world models.

\subsection{Joint Camera--Hand Conditioning}
\label{sec:joint_conditioning}

\textbf{Camera conditioning.}
We build on the LTX-2.5~\cite{hacohen2026ltx2efficientjointaudiovisual} image-to-video backbone with a SolarWM camera adapter. Camera intrinsics and relative extrinsics are incorporated into self-attention through PRoPE. We retain this camera-conditioning pathway while introducing two complementary branches for hand control.

\textbf{Hand conditioning.}
The spatial branch encodes rendered skeleton videos $S$ with the video VAE~\cite{kingma2013auto} and projects their latents into the video-token space. The motion branch encodes numerical hand features $H$, consisting of wrist translation, joint geometry, and a validity mask, using a temporal encoder $E_m$ with linear projections and a residual temporal convolution. Given noisy video latents $z_\tau$, we form the Transformer input as
\begin{equation}
\begin{aligned}
U ={}& W_z P(z_\tau)
     + W_s P\!\left(E_{\mathrm{VAE}}(S)\right)
     + \mathcal{B}\!\left(E_m(H)\right),
\end{aligned}
\label{eq:hand_injection}
\end{equation}
where $P$ denotes patchification, $W_z$ and $W_s$ are the video and skeleton projections, and $\mathcal{B}$ broadcasts each motion embedding across the spatial tokens at the corresponding latent time position. The two branches provide spatially localized skeleton features and temporally contextualized motion features.

\textbf{Joint adaptation.}
We freeze the pretrained backbone and camera LoRA\cite{hu2022lora}, and jointly optimize an additional hand LoRA with the two hand-conditioning branches. The skeleton projection and the motion encoder's output projection are bias-free and zero-initialized, so the added conditioning residual is initially zero.

To emphasize hand reconstruction while retaining
scene-wide supervision, we use a weighted denoising loss:
\begin{equation}
\mathcal{L}_{\mathrm{joint}}
=
\mathbb{E}\!\left[
\frac{\sum_i w_i\|\hat{y}_i-y_i\|_2^2}
     {\sum_i w_i}
\right],
\quad
w_i=m_i(1+\lambda_{\mathrm{hand}}r_i),
\label{eq:hand_weighted_loss}
\end{equation}

where $\hat{y}_i$ and $y_i$ are the prediction and denoising target for video token $i$, respectively.
The binary mask $r_i$ identifies hand-region tokens, $m_i$ excludes the initial conditioning latent, and $\lambda_{\mathrm{hand}}$ controls the additional hand weight. Setting $\lambda_{\mathrm{hand}}=0$ recovers uniform weighting over valid tokens.

\subsection{Causal Adaptation and Asymmetric DMD}
\label{sec:causal_distillation}

We obtain a streaming generator in two stages: teacher-forced causal adaptation, followed by asymmetric Distribution Matching Distillation (DMD)~\cite{Yin_2024_CVPR} from a bidirectional teacher  to obtain a few-step causal generator.

\textbf{Teacher-forced causal adaptation.}
We initialize from the bidirectional joint-control teacher, freeze the backbone, camera and hand LoRAs, and both hand-conditioning branches, and train only an additional causal-adaptation LoRA.
We adopt bidirectional attention within each chunk and causal attention across chunks.
Each chunk is jointly denoised with access to its own tokens and the available preceding chunks. During teacher-forced training, preceding video chunks are supplied from ground-truth sequences. We also use left-padded temporal convolutions in the motion encoder and restrict skeleton conditions to the available temporal range to prevent information leakage from future chunks.

\textbf{Asymmetric DMD with self-forcing.}
We initialize the few-step student from the causally adapted model and distill it using the frozen bidirectional joint-control teacher. During distillation, the student performs self-forcing~\cite{huang2026self} rollouts, conditioning each chunk on its own generated history:
\begin{equation}
\hat{z}^{(k)}=G_\theta\!\left(\xi^{(k)};\hat{z}^{(<k)},c_{\leq k}\right),
\label{eq:self_forcing_rollout}
\end{equation}
where $G_\theta$ denotes the complete few-step sampling procedure, $\xi^{(k)}$ denotes its sampling noise, and $c_{\leq k}$ contains the initial image, scene-description prompt, and camera--hand controls available when generating chunk $k$.

We apply Distribution Matching Distillation (DMD) to align the student and teacher distributions through a noise-averaged reverse KL objective. Given the generated sequence $\hat{z}$, we construct $\hat{z}_\tau=\alpha_\tau\hat{z}+\sigma_\tau\epsilon$, where $\epsilon\sim\mathcal{N}(0,I)$. A trainable fake-score network $s_\psi$ estimates the student distribution score, while the frozen teacher provides $s_T$. Their difference yields the generator update:
\begin{equation}
\begin{aligned}
\nabla_\theta\mathcal{L}_{\mathrm{DMD}}\approx \\ 
&\mathbb{E}\Big[w(\tau)J_\theta^\top\cdot\big(s_\psi(\hat{z}_\tau,\tau;c)-s_T(\hat{z}_\tau,\tau;c)\big)\Big],
\end{aligned}
\label{eq:dmd_gradient}
\end{equation}
where $J_\theta=\partial\hat{z}/\partial\theta$, $c$ denotes the conditioning information for the full distillation sequence, and $w(\tau)$ incorporates the noise weighting and the perturbation factor $\alpha_\tau$. Here, $s_\psi$ and $s_T$ denote the student-distribution and teacher-distribution scores, respectively, expressed in a common score parameterization.

\paragraph{Online inference.}
The distilled student generates video sequentially using four denoising steps per chunk. Each chunk conditions on the initial frame, a bounded history of generated chunks, and the available camera and hand controls. Committed outputs remain unchanged. The interface continues to receive updated controls during generation and incorporates them into subsequent chunks, allowing users to adjust their actions based on visual feedback.

\begin{figure*}[t]
  \centering
\includegraphics[width=1.0\linewidth]{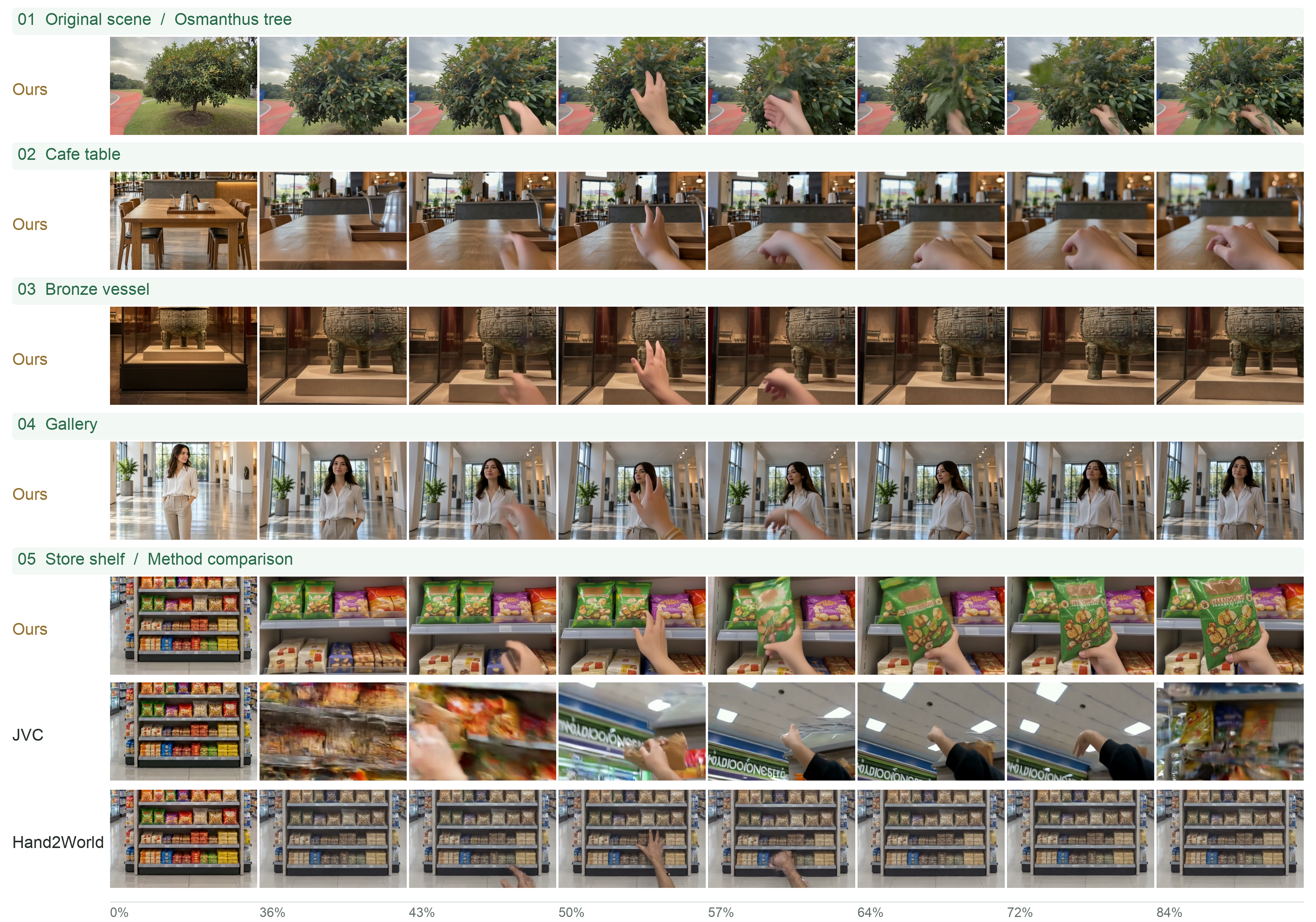}
    \caption{Generation across scenes using shared camera and hand controls from an osmanthus-tree interaction. The final case compares methods on the store-shelf scene. Frames are sampled at matched relative progress.}
   \label{fig:maincase}
\end{figure*}

\begin{figure*}[t]
  \centering
\includegraphics[width=1.0\linewidth]{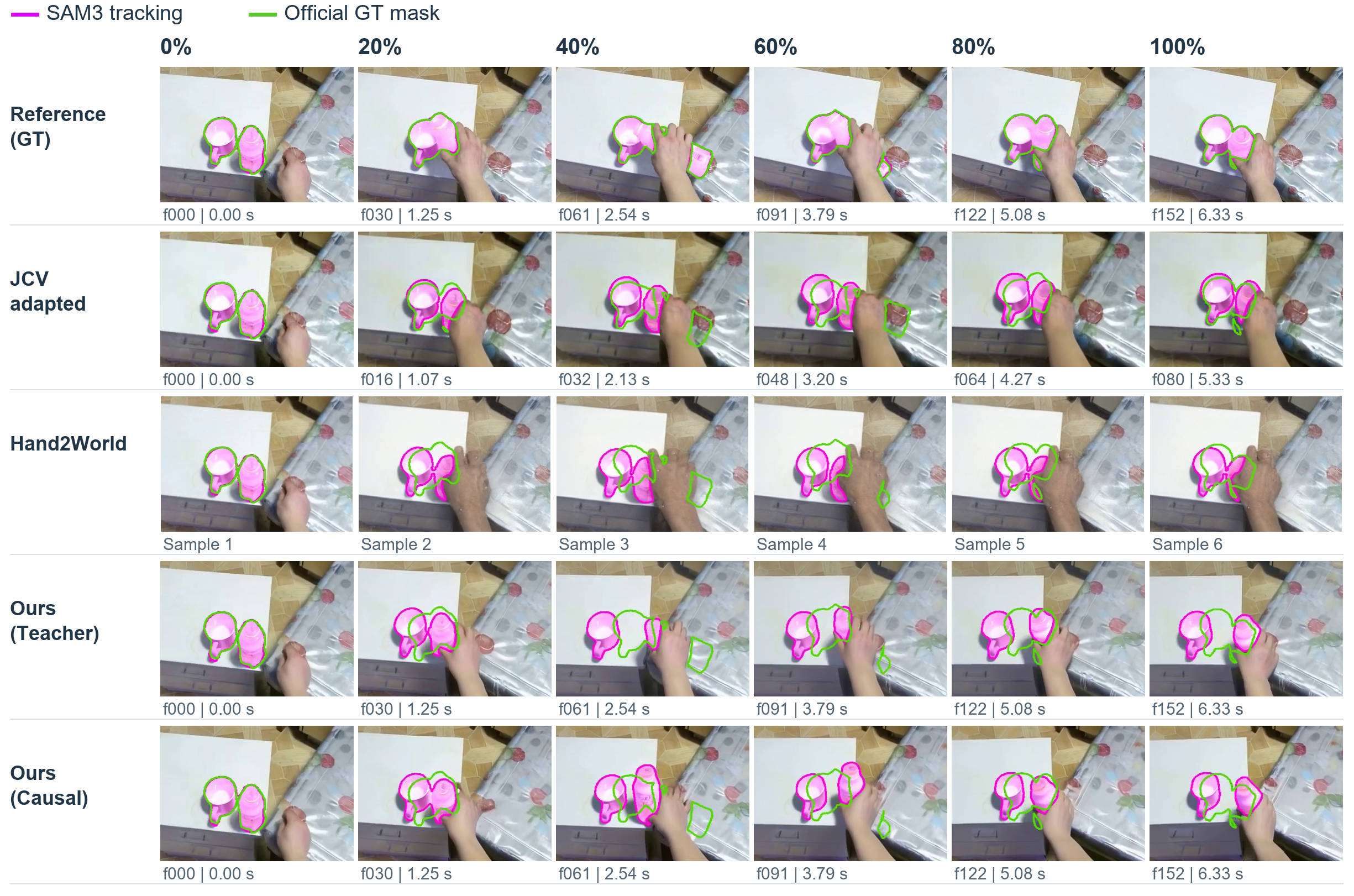}
\caption{Object-response comparison on HOI4D. Magenta overlays show SAM3-tracked masks; green contours show reference masks. Image-space overlap can differ from action-consistent object motion.}
   \label{fig:bottle}
\end{figure*}

\begin{figure*}[t]
  \centering
\includegraphics[width=1.0\linewidth]{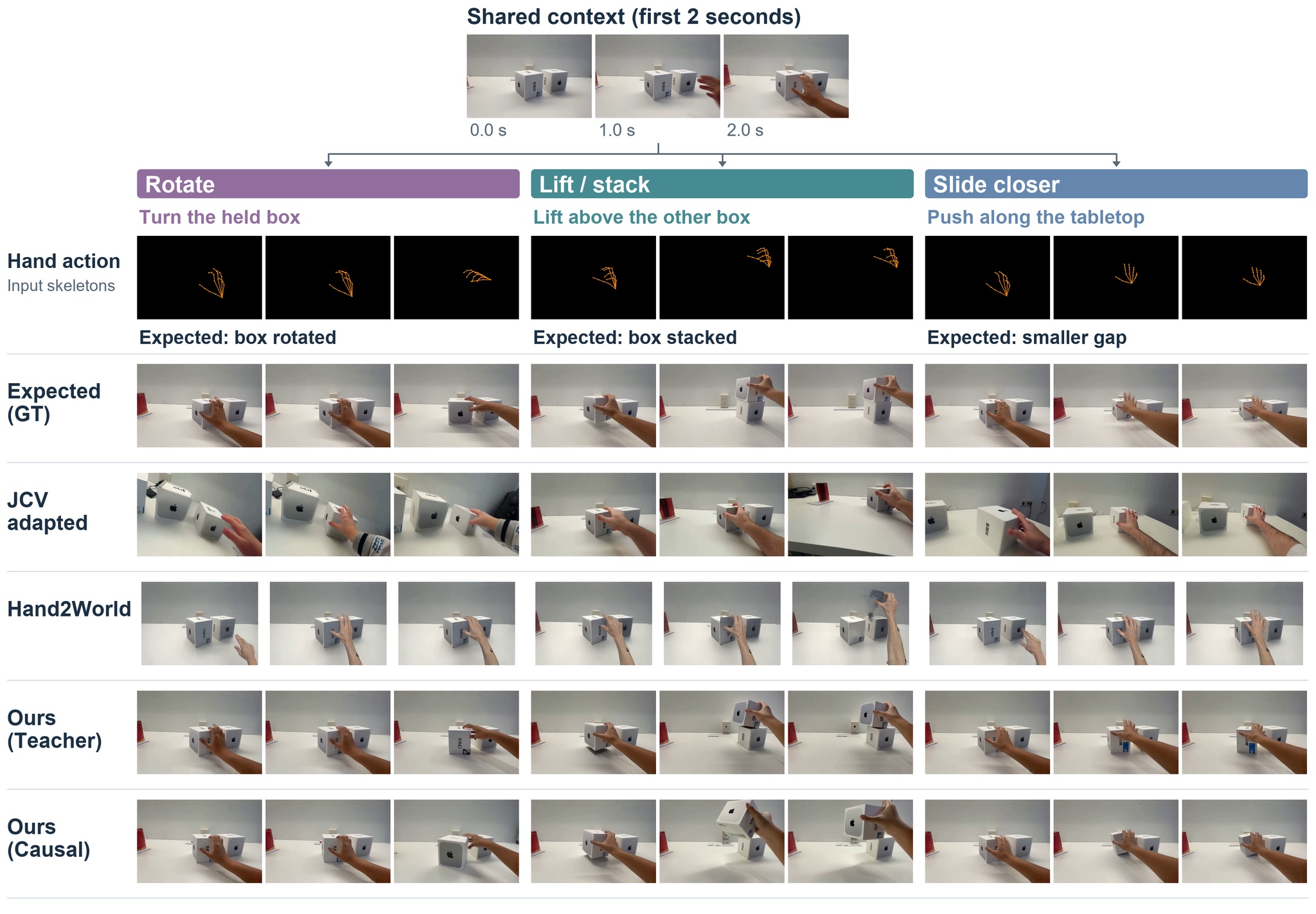}
\caption{Action-dependent continuations from a shared two-second source context under rotating, lifting/stacking, and sliding controls. Branch labels describe evaluation goals and are not supplied as action-text prompts.}
   \label{fig:three}
\end{figure*}

\subsection{Touch\'{e}Bench}
\label{sec:handworldbench}

We introduce Touch\'{e}Bench to evaluate four aspects of camera--hand-controlled world generation: hand-motion following, camera--hand coordination, online interaction, and object response (Fig.~\ref{fig:bench}). The benchmark combines reference-based evaluation on 200 held-out HOI4D clips with purpose-recorded scenarios for joint camera--hand control and action-dependent continuations. Hand-motion following and object response are evaluated on the same HOI4D clips and generated outputs, allowing control accuracy and object-response fidelity to be assessed together.

\paragraph{Q1: Hand-motion following.}
We condition generation on the hand trajectories associated with each HOI4D clip and estimate hand keypoints from the generated frames. Following temporal alignment and hand correspondence matching, we compare detected joints $\hat{\mathbf{u}}_{t,j}$ with reference joints $\mathbf{u}_{t,j}$ in a common normalized image coordinate system. We report the percentage of correct keypoints:
\begin{equation}
\operatorname{PCK}_{\delta}
=
\frac{1}{|\mathcal{V}|}
\sum_{(t,j)\in\mathcal{V}}
\mathbb{1}\!\left[
d_{t,j}=1
\;\land\;
\|\hat{\mathbf{u}}_{t,j}-\mathbf{u}_{t,j}\|_2
\leq \delta
\right],
\label{eq:bench_pck}
\end{equation}
where $\mathcal{V}$ contains valid reference joints, $d_{t,j}$ indicates a matched detection, and $\delta$ is the normalized distance threshold. We report PCK at $\delta=0.2$, counting missing detections as incorrect. Joint localization error on matched detections and detection coverage are reported as supplementary measures. Scores are computed per clip and averaged across clips; the initial conditioning frame is excluded.

\paragraph{Q2: Camera--hand coordination.}
We evaluate joint camera--hand control on ten purpose-recorded clips combining camera movement with hand interaction. Each clip uses first-frame image-to-video initialization and the camera and hand inputs supported by each method; JCV receives hand conditioning without camera control. GPT-6 Astra \cite{openai2026gpt6}, accessed through Codex \cite{openai2026codex}, assesses each continuation using six binary checklist questions covering visual consistency, viewpoint attainment, interaction execution, and the final object state. We report Camera as the viewpoint-attainment pass rate and Interaction as the average pass rate of interaction-execution and final-state judgments. Scores are averaged equally across clips.

\paragraph{Q3: Online interaction.}
We evaluate action-dependent continuations using eight purpose-recorded scenarios, each containing three alternative hand-action branches following a shared approximately two-second context. The scene-description prompt remains unchanged across branches and does not specify the target action. Each branch supplies its corresponding hand controls to test whether the generated continuation realizes the intended interaction and distinguishes it from alternative outcomes. GPT-6 Astra \cite{openai2026gpt6}, accessed through Codex \cite{openai2026codex}, assesses the 24 continuations using six binary checklist questions per continuation, yielding 144 judgments. Three basic criteria assess visual and scene consistency, while three hard criteria assess action execution ($A$), the intended final state ($S$), and exclusion of alternative outcomes ($X$). We report the mean pass rates for Basic, Hard, and Overall.

\paragraph{Q4: Object response.}
Using the same HOI4D-200 evaluation set, we compare target-object responses in generated and reference videos. We obtain temporally corresponding object masks with SAM3~\cite{carion2026sam} and evaluate mask overlap and centroid displacement:
\begin{equation}
\begin{aligned}
\operatorname{IoU}_t
&=
\frac{|\hat{M}_t\cap M_t|}
     {|\hat{M}_t\cup M_t|},\\
E_{\mathrm{center},t}
&=
\|\mathbf{c}(\hat{M}_t)-\mathbf{c}(M_t)\|_2,
\end{aligned}
\label{eq:bench_object}
\end{equation}
where $\hat{M}_t$ and $M_t$ denote generated and reference object masks, respectively, and $\mathbf{c}(\cdot)$ denotes the mask centroid in image coordinates. We report mask IoU and centroid displacement in pixels. Frame-level measurements are averaged within each clip and then equally across clips, excluding the initial conditioning frame. Target-tracking coverage is reported separately.

\begin{table*}
\centering
\caption{Video quality on unseen in-domain scenes and out-of-domain HOT3D. All our variants use hand-weighted loss. Best results are bold; KID is unscaled.}
\label{tab:video_quality}
\setlength{\tabcolsep}{3pt}
\renewcommand{\arraystretch}{1.15}
\resizebox{\textwidth}{!}{%
\begin{tabular}{@{}l*{15}{r}@{}}
\toprule
& \multicolumn{5}{c}{EgoStandard-200} & \multicolumn{5}{c}{EgoLife-300} & \multicolumn{5}{c}{HOT3D-45 (OOD)} \\
\cmidrule(lr){2-6}\cmidrule(lr){7-11}\cmidrule(l){12-16}
Method & PSNR$\uparrow$ & SSIM$\uparrow$ & LPIPS$\downarrow$ & DINOv2$\uparrow$ & KID$\downarrow$ & PSNR$\uparrow$ & SSIM$\uparrow$ & LPIPS$\downarrow$ & DINOv2$\uparrow$ & KID$\downarrow$ & PSNR$\uparrow$ & SSIM$\uparrow$ & LPIPS$\downarrow$ & DINOv2$\uparrow$ & KID$\downarrow$ \\
\midrule
\multicolumn{15}{l}{\textit{Bidirectional}} \\
LTX-2.5 I2V & 15.95 & 0.615 & 0.551 & 0.724 & 0.00887 & 13.61 & 0.473 & 0.635 & 0.537 & \textbf{0.00648} & 14.04 & 0.517 & 0.558 & 0.707 & 0.01766 \\
SolarWM (camera-only) & 16.36 & 0.643 & 0.494 & 0.728 & 0.01732 & 14.53 & 0.527 & 0.584 & 0.542 & 0.01033 & 14.77 & 0.543 & 0.513 & 0.719 & 0.01154 \\
JCV & 16.63 & 0.643 & 0.486 & 0.757 & 0.00308 & 13.47 & 0.473 & 0.647 & 0.486 & 0.01533 & 12.64 & 0.470 & 0.634 & 0.726 & 0.02330 \\
Hand2World & 13.81 & 0.619 & 0.635 & 0.402 & 0.07425 & 12.08 & 0.468 & 0.692 & 0.354 & 0.02642 & 13.63 & 0.512 & 0.553 & 0.600 & 0.07396 \\
Ours (bidirectional teacher) & \textbf{19.18} & \textbf{0.708} & \textbf{0.345} & \textbf{0.816} & \textbf{0.00265} & \textbf{16.19} & 0.572 & \textbf{0.488} & \textbf{0.652} & 0.00657 & 15.11 & 0.552 & \textbf{0.488} & \textbf{0.769} & \textbf{0.00617} \\
\midrule
\multicolumn{15}{l}{\textit{Causal}} \\
Ours (causal) & 17.60 & 0.666 & 0.410 & 0.785 & 0.00367 & 15.05 & 0.519 & 0.556 & 0.582 & 0.00876 & 14.74 & 0.525 & 0.500 & 0.750 & 0.00747 \\
Ours (causal-4 step) & 18.38 & 0.696 & 0.408 & 0.742 & 0.00852 & 16.04 & \textbf{0.578} & 0.589 & 0.437 & 0.00746 & \textbf{15.17} & \textbf{0.556} & 0.499 & 0.732 & 0.01180 \\

\bottomrule
\end{tabular}%
}
\end{table*}

\section{Experiments}

\subsection{Experimental Setup}
\label{sec:experimental_setup}
\paragraph{Datasets.}
We train our model on 40,958 egocentric video clips, comprising 24,000 EgoLife~\cite{Yang_2025_CVPR} clips, 15,918 EgoStandard clips, and 1,040 auxiliary clips. The auxiliary set includes hand-emergence clips, in which hands are removed from the initial frame and appear in subsequent chunks, as well as static clips with and without hands. We evaluate on EgoStandard-200 \cite{lightwheel2026egosuite}, EgoLife-300 \cite{Yang_2025_CVPR}, and HOT3D-45 \cite{Banerjee_2025_CVPR}. All three test sets contain scenes unseen during training. EgoStandard and EgoLife evaluate generalization to unseen scenes within the training domains, while HOT3D provides an out-of-domain evaluation. For HOT3D, we follow the original 81-frame configuration.

\paragraph{Baselines.}
Few existing methods closely match our joint camera--hand control setting. We compare against the following models:
\begin{enumerate}
    \item LTX-2.5 I2V and the camera-only SolarWM model, which serve as baselines for video generation quality.
    \item JCV~\cite{zhang2026controllable} and Hand2World~\cite{wang2026hand2worldautoregressiveegocentricinteraction}, two related hand-conditioned video generation models with publicly available weights. We use their released weights without additional training or fine-tuning, adapting the input representation, resolution, and evaluation interface to our test protocol. Each method receives the inputs supported by its original implementation. In particular, JCV does not expose a camera-control interface and therefore receives hand conditioning only. For Hand2World, we project 3D MANO meshes to construct the hand-conditioning input used by its original pipeline. We use the same evaluation examples and scoring implementation across methods, with method-specific context initialization and temporal alignment described below.
\end{enumerate}

We report our bidirectional teacher, the multi-step causal model obtained through teacher-forced adaptation, and the four-step causal model obtained through self-forcing DMD to examine video quality and interaction performance across training stages.

\paragraph{Evaluation protocol.}
We evaluate video quality using PSNR, SSIM~\cite{wang2004ssim}, LPIPS~\cite{zhang2018lpips}, DINOv2~\cite{oquab2024dinov2} similarity, and KID~\cite{binkowski2018kid}, excluding the initial conditioning frame. The four paired metrics are averaged within each clip and then equally across clips, while KID measures distributional similarity over sampled frames. All methods are evaluated using a shared scoring implementation. We additionally assess hand-motion following, camera--hand coordination, online interaction, and object response using Touch\'{e}Bench.

\subsection{Video Generation Quality}
\label{sec:video_quality}

Table~\ref{tab:video_quality} compares video generation quality across two in-domain test sets with unseen scenes and the out-of-domain HOT3D-45 benchmark. Our bidirectional teacher achieves strong results across all three datasets. Relative to camera-only SolarWM, it improves PSNR by 2.82 dB on EgoStandard-200 and 1.66 dB on EgoLife-300, while reducing LPIPS from 0.494 to 0.345 and from 0.584 to 0.488, respectively. On HOT3D-45, it achieves an LPIPS of 0.488 and a DINOv2 similarity of 0.769. These results show improved agreement with reference observations across both in-domain and out-of-domain scenes. Figure~\ref{fig:maincase} further illustrates the visual results across diverse scenes, where the same camera and hand controls are reused with different initial images and scene descriptions.

Our multi-step causal model also achieves lower LPIPS and higher DINOv2 similarity than SolarWM on all three datasets, while its PSNR improves or remains comparable. The four-step distilled model further improves PSNR and SSIM over the multi-step causal model across all three datasets, with mixed changes in LPIPS, DINOv2 similarity, and KID. These results characterize the visual-quality trade-offs introduced by causal adaptation and few-step distillation.

We next evaluate hand-motion following, camera--hand coordination, online interaction, and object response using Touch\'{e}Bench to examine how these models translate visual generation quality into controllable interaction.

\subsection{Evaluation on Touch\'{e}Bench}
\label{sec:interaction_results}

\begin{table*}[t]
\centering
\caption{Touch\'{e}Bench results. All our variants use hand-weighted loss. PCK, checklist scores, and IoU are percentages; L2 and target-center offset (TO) are in pixels. Q2 Interaction averages action-execution and final-state pass rates. Best results are bold.}
\label{tab:handworldbench}
\setlength{\tabcolsep}{5pt}
\renewcommand{\arraystretch}{1.12}
\resizebox{\textwidth}{!}{%
\begin{tabular}{lccccccccc}
\toprule
\multirow{2}{*}{Method}
& \multicolumn{2}{c}{Q1: Hand following}
& \multicolumn{2}{c}{Q2: Camera--hand coordination}
& \multicolumn{3}{c}{Q3: Online interaction}
& \multicolumn{2}{c}{Q4: Object response} \\
\cmidrule(lr){2-3}
\cmidrule(lr){4-5}
\cmidrule(lr){6-8}
\cmidrule(lr){9-10}
& PCK@0.2 $\uparrow$
& L2 $\downarrow$
& Camera $\uparrow$
& Interaction $\uparrow$
& Basic $\uparrow$
& Hard $\uparrow$
& Overall $\uparrow$
& IoU $\uparrow$
& TO $\downarrow$ \\
\midrule
JCV
& 59.1 & 34.3
& 50.0 & 15.0
& 79.2 & 23.6 & 51.4
& \textbf{38.7} & 54.0 \\
Hand2World
& 61.2 & \textbf{28.2}
& 40.0 & 15.0
& 48.6 & 18.1 & 33.3 &
28.6 & 189.3 \\
\midrule
Ours (Teacher)
& \textbf{65.8} & 48.2
& \textbf{100.0} & \textbf{60.0}
& \textbf{97.2} & 45.8 & 71.5
& 36.6 & 49.3 \\
Ours (Causal)
& 57.9 & 46.8
& 70.0 & 40.0
& 94.4 & \textbf{55.6} & \textbf{75.0}
& 34.9 & \textbf{48.8} \\
Ours (Causal-4 step)
& 54.0 & 58.9
& 60.0 & 35.0
& 93.1 & 44.4 & 68.8
& 35.0 & 49.3 \\
\bottomrule
\end{tabular}%
}
\end{table*}

Table~\ref{tab:handworldbench} reports results on the four dimensions of Touch\'{e}Bench. Our bidirectional teacher achieves the highest hand-following PCK@0.2 of 65.8\%, compared with 61.2\% for Hand2World and 59.1\% for JCV, while our multi-step and four-step causal models obtain 57.9\% and 54.0\%, respectively. Hand2World achieves the lowest localization error on matched detections at 28.2 pixels. For camera--hand coordination, our teacher achieves 100.0\% Camera and 60.0\% Interaction pass rates, compared with 50.0\% and 15.0\% for JCV, and 40.0\% and 15.0\% for Hand2World. Our causal models retain this advantage, with the multi-step model scoring 70.0\% and 40.0\%, and the four-step model scoring 60.0\% and 35.0\%. On online interaction, the multi-step causal model achieves the highest overall pass rate of 75.0\% and hard-criterion pass rate of 55.6\%, exceeding the teacher's 71.5\% and 45.8\%, respectively. The four-step model obtains 68.8\% overall, remaining above JCV's 51.4\% and Hand2World's 33.3\%. For object response, JCV achieves the highest mask IoU of 38.7\%, while our multi-step causal model achieves the lowest target-center offset of 48.8 pixels. These results highlight the distinction between image-space alignment and interaction success: favorable keypoint or object-overlap scores can coexist with low action-execution and final-state pass rates. The qualitative case study \ref{fig:bottle} illustrates this distinction through examples of camera movement and object behavior, while the action-branch examples \ref{fig:three} show different continuations from a shared context. Together, the quantitative and qualitative results demonstrate stronger joint-control and action-dependent interaction performance for our framework, with reduced hand-following and camera--hand coordination scores after causal adaptation and few-step distillation.

\begin{table*}[t]
\centering
\caption{Effect of hand-weighted loss on video quality. Better results within each Teacher or Causal pair are bold, including ties.}
\label{tab:ablation_hand_loss_quality}
\setlength{\tabcolsep}{9pt}
\renewcommand{\arraystretch}{1.08}
\begin{tabular}{llccccc}
\toprule
Model & Hand loss & PSNR $\uparrow$ & SSIM $\uparrow$ & LPIPS $\downarrow$ & DINO $\uparrow$ & KID $\downarrow$ \\
\midrule
\multicolumn{7}{l}{\textit{EgoStandard-200}} \\
Teacher & w/o & 19.13 & 0.705 & 0.350 & 0.812 & 0.00280 \\
Teacher & w/ & \textbf{19.18} & \textbf{0.708} & \textbf{0.345} & \textbf{0.816} & \textbf{0.00265} \\
\addlinespace[2pt]
Causal & w/o & \textbf{17.61} & \textbf{0.668} & 0.416 & 0.780 & \textbf{0.00354} \\
Causal & w/ & 17.60 & 0.666 & \textbf{0.410} & \textbf{0.785} & 0.00367 \\
\midrule
\multicolumn{7}{l}{\textit{EgoLife-300}} \\
Teacher & w/o & 15.97 & 0.564 & 0.503 & 0.633 & \textbf{0.00630} \\
Teacher & w/ & \textbf{16.19} & \textbf{0.572} & \textbf{0.488} & \textbf{0.652} & 0.00657 \\
\addlinespace[2pt]
Causal & w/o & 14.85 & 0.516 & 0.568 & 0.558 & 0.01215 \\
Causal & w/ & \textbf{15.05} & \textbf{0.519} & \textbf{0.556} & \textbf{0.582} & \textbf{0.00876} \\
\midrule
\multicolumn{7}{l}{\textit{HOT3D-45}} \\
Teacher & w/o & \textbf{15.14} & \textbf{0.552} & 0.489 & 0.759 & 0.00704 \\
Teacher & w/ & 15.11 & \textbf{0.552} & \textbf{0.488} & \textbf{0.769} & \textbf{0.00617} \\
\addlinespace[2pt]
Causal & w/o & \textbf{14.77} & 0.515 & \textbf{0.500} & 0.743 & 0.01005 \\
Causal & w/ & 14.74 & \textbf{0.525} & \textbf{0.500} & \textbf{0.750} & \textbf{0.00747} \\
\midrule
\multicolumn{7}{l}{\textit{HOI4D-200}} \\
Teacher & w/o & 15.74 & 0.663 & \textbf{0.438} & \textbf{0.792} & \textbf{0.00210} \\
Teacher & w/ & \textbf{15.76} & \textbf{0.665} & 0.441 & \textbf{0.792} & 0.00240 \\
\addlinespace[2pt]
Causal & w/o & \textbf{15.55} & \textbf{0.660} & \textbf{0.445} & \textbf{0.787} & 0.00260 \\
Causal & w/ & 15.39 & 0.656 & 0.450 & 0.786 & \textbf{0.00248} \\
\bottomrule
\end{tabular}
\end{table*}

\subsection{Case Study}
\label{sec:case_study}

Figure~\ref{fig:maincase} illustrates how a shared camera--hand control sequence produces scene-dependent interactions. Starting from the osmanthus-tree example, we reuse the controls with four different scene initializations. The hand settles on the tabletop in the cafe scene, moves out of view after reaching through free space in the gallery, and grasps a package at the store shelf. These examples show that the generated hand behavior adapts to the surrounding scene while retaining the input motion pattern. In the shelf comparison, JCV produces substantial viewpoint changes toward the ceiling, while Hand2World generates hand motion in front of a largely unchanged shelf; our model instead produces a reach-and-lift interaction. Figure~\ref{fig:bottle} further examines the distinction between spatial agreement and object interaction. In the reference, the hand moves the bottle relative to the cup. JCV retains the bottle near its initial position through much of the sequence, whereas our teacher and causal model generate more pronounced hand-associated object movement. Retaining an object near its initial location can preserve overlap with reference masks over parts of the sequence, illustrating why mask IoU should be interpreted alongside action execution and object-state changes. Finally, Figure~\ref{fig:three} evaluates action-dependent continuations from a shared approximately two-second context. Different hand inputs ask the model to rotate a box, lift it above another box, or slide it along the tabletop. Our teacher and causal model produce distinct changes in box orientation, height, and position across these branches, while the baselines show less consistent separation of the intended outcomes. Together, these cases illustrate joint camera--hand control, scene-dependent contact behavior, and distinct object responses to actions introduced after a shared context.


\begin{table*}[t]
\centering
\caption{Effect of hand-weighted loss on Touch\'{e}Bench. Scores are percentages except L2 (pixels). Weighted variants match the main-table models. Better results within each model pair are bold, including ties.}
\label{tab:ablation_hand_loss_bench}
\setlength{\tabcolsep}{5pt}
\renewcommand{\arraystretch}{1.0}
\resizebox{\textwidth}{!}{%
\begin{tabular}{llcccccccc}
\toprule
\multirow{2}{*}{Model} & \multirow{2}{*}{Hand loss}
& \multicolumn{2}{c}{Q1: Hand following}
& \multicolumn{2}{c}{Q2: Camera--hand coordination}
& \multicolumn{2}{c}{Q3: Online interaction}
& \multicolumn{2}{c}{Q4: Object response} \\
\cmidrule(lr){3-4}\cmidrule(lr){5-6}\cmidrule(lr){7-8}\cmidrule(lr){9-10}
& & PCK@0.2 $\uparrow$ & L2 $\downarrow$
& Camera $\uparrow$ & Interaction $\uparrow$
& Overall $\uparrow$ & Hard $\uparrow$
& BBox IoU $\uparrow$ & Mask IoU $\uparrow$ \\
\midrule
Teacher & w/o & 58.7 & 60.0 & \textbf{100.0} & 40.0 & 65.3 & 37.5 & 68.4 & 32.5 \\
Teacher & w/ & \textbf{65.8} & \textbf{48.2} & \textbf{100.0} & \textbf{60.0} & \textbf{71.5} & \textbf{45.8} & \textbf{69.5} & \textbf{36.6} \\
\midrule
Causal & w/o & 55.0 & 55.0 & \textbf{90.0} & \textbf{45.0} & 72.2 & 50.0 & 68.5 & 33.6 \\
Causal & w/ & \textbf{57.9} & \textbf{46.8} & 70.0 & 40.0 & \textbf{75.0} & \textbf{55.6} & \textbf{69.0} & \textbf{34.9} \\
\bottomrule
\end{tabular}%
}
\end{table*}

\subsection{Ablation Studies and Analysis}
\label{sec:ablation}

\paragraph{Hand-weighted loss.}
Tables~\ref{tab:ablation_hand_loss_quality}, \ref{tab:ablation_hand_loss_details}, and \ref{tab:ablation_hand_loss_bench} examine the effect of hand-weighted training on video quality and interaction. Hand weighting improves keypoint accuracy for both the teacher and causal models across all three PCK thresholds. For the teacher, PCK@0.2 increases from 58.7\% to 65.8\%, while localization error decreases from 60.0 to 48.2 pixels. The causal model improves from 55.0\% to 57.9\% PCK, with localization error decreasing from 55.0 to 46.8 pixels. These improvements also extend to online interaction: Q3 Overall increases from 65.3\% to 71.5\% for the teacher and from 72.2\% to 75.0\% for the causal model. For camera--hand coordination, the teacher's Interaction score increases from 40.0\% to 60.0\%, while its Camera score remains at 100.0\%. The causal model exhibits a trade-off, with Camera decreasing from 90.0\% to 70.0\% and Interaction from 45.0\% to 40.0\%. Object-mask IoU improves for both model types. Changes in global video quality are dataset-dependent: all five metrics improve for the teacher on EgoStandard and for the causal model on EgoLife, while the remaining evaluation sets show mixed changes. Overall, hand-weighted supervision strengthens hand following and online interaction, with a coordination trade-off in the causal model.

\paragraph{Causal adaptation and few-step distillation.}
We examine the generation stages through the teacher, multi-step causal, and four-step causal results in Tables~\ref{tab:video_quality} and \ref{tab:handworldbench}. Moving from the bidirectional teacher to the causal model reduces reference-based video quality and hand-following accuracy, with PCK@0.2 decreasing from 65.8\% to 57.9\%. Q2 Camera and Interaction also decrease from 100.0\% and 60.0\% to 70.0\% and 40.0\%, respectively. On Q3, the causal model achieves higher Overall and Hard pass rates, increasing from 71.5\% and 45.8\% to 75.0\% and 55.6\% under the method-specific context initialization used in our evaluation. This contrast shows that stronger reference reconstruction and keypoint alignment do not directly imply better action-dependent continuations. Four-step distillation improves PSNR and SSIM over the multi-step causal model on all three video-quality test sets, while changes in LPIPS, DINOv2 similarity, and KID are mixed. Hand-following PCK decreases to 54.0\%, Q2 Camera and Interaction decrease to 60.0\% and 35.0\%, and Q3 Overall decreases to 68.8\%. Object-response measurements remain close, with mask IoU changing from 34.9\% to 35.0\% and target-center offset from 48.8 to 49.3 pixels. These results characterize the transition to four-step causal generation: object-response fidelity remains similar under the reported spatial metrics, while hand following, joint control, and online interaction show larger performance reductions.


\begin{table*}[t]
\centering
\caption{Detailed hand-loss ablation. Scores are percentages except L2 (pixels). $C$, $A$, $S$, and $X$ denote viewpoint attainment, action execution, final-state correctness, and alternative-outcome exclusion. Better results within each model pair are bold, including ties.}
\label{tab:ablation_hand_loss_details}
\setlength{\tabcolsep}{8pt}
\renewcommand{\arraystretch}{0.9}
\textbf{(a) Q1: Hand-motion following on HOI4D-200}\\[3pt]
\begin{tabular}{llcccc}
\toprule
Model & Hand loss & PCK@0.05 $\uparrow$ & PCK@0.10 $\uparrow$ & PCK@0.20 $\uparrow$ & L2 (px) $\downarrow$ \\
\midrule
Teacher & w/o & 10.0 & 29.3 & 58.7 & 60.0 \\
Teacher & w/ & \textbf{13.7} & \textbf{36.6} & \textbf{65.8} & \textbf{48.2} \\
\midrule
Causal & w/o & 8.8 & 26.2 & 55.0 & 55.0 \\
Causal & w/ & \textbf{10.1} & \textbf{28.9} & \textbf{57.9} & \textbf{46.8} \\
\bottomrule
\end{tabular}
\par\vspace{8pt}
\textbf{(b) Q2: Camera--hand coordination}\\[3pt]
\begin{tabular}{llccccccc}
\toprule
Model & Hand loss & Overall & Basic & Hard & Camera ($C$) & Interaction & $A$ & $S$ \\
\midrule
Teacher & w/o & 80.0 & \textbf{100.0} & 60.0 & \textbf{100.0} & 40.0 & 40.0 & 40.0 \\
Teacher & w/ & \textbf{86.7} & \textbf{100.0} & \textbf{73.3} & \textbf{100.0} & \textbf{60.0} & \textbf{60.0} & \textbf{60.0} \\
\midrule
Causal & w/o & \textbf{80.0} & \textbf{100.0} & \textbf{60.0} & \textbf{90.0} & \textbf{45.0} & \textbf{40.0} & \textbf{50.0} \\
Causal & w/ & 68.3 & 86.7 & 50.0 & 70.0 & 40.0 & \textbf{40.0} & 40.0 \\
\bottomrule
\end{tabular}
\par\vspace{8pt}
\textbf{(c) Q3: Action-dependent online interaction}\\[3pt]
\begin{tabular}{llcccccc}
\toprule
Model & Hand loss & Overall & Basic & Hard & $S$ & $X$ & $A$ \\
\midrule
Teacher & w/o & 65.3 & 93.1 & 37.5 & 29.2 & 58.3 & 25.0 \\
Teacher & w/ & \textbf{71.5} & \textbf{97.2} & \textbf{45.8} & \textbf{37.5} & \textbf{66.7} & \textbf{33.3} \\
\midrule
Causal & w/o & 72.2 & \textbf{94.4} & 50.0 & \textbf{41.7} & \textbf{66.7} & \textbf{41.7} \\
Causal & w/ & \textbf{75.0} & \textbf{94.4} & \textbf{55.6} & 33.3 & \textbf{66.7} & 33.3 \\
\bottomrule
\end{tabular}
\par\vspace{3pt}
\begin{minipage}{0.97\textwidth}
\footnotesize All checklist columns are higher-is-better. Q3 causal models use an approximately two-second context; teachers are initialized from the context endpoint. Loss effects are compared within each model type.
\end{minipage}
\end{table*}

\section{Conclusion and Future Work}
\label{sec:conclusion}

We presented Touch\'{e}World, a video world modeling framework that integrates continuous hand interaction with camera-guided exploration. We constructed joint camera--hand training data through annotation completion, consistency-based filtering, and temporal alignment, and combined spatial skeleton conditioning with numerical hand-motion features through parameter-efficient adaptation. Causal adaptation and few-step distillation enable chunk-level online generation with updated camera and hand controls. We further introduced Touch\'{e}Bench to evaluate hand-motion following, camera--hand coordination, online interaction, and object response. Experiments demonstrate scene-dependent interaction across diverse environments and action-dependent continuations from shared contexts, while revealing distinct trade-offs among visual quality, hand following, and interaction performance.

Our qualitative results also reveal early signs of gesture-mediated interaction with people in generated scenes. This opens a direction for extending interactive world models beyond object manipulation toward human interaction. Future work will explore how generated people perceive and respond to user gestures, enabling richer exchanges between users and the inhabitants of generative worlds.

\section{Acknowledgements}
This work is partially supported by a public grant overseen by the French National Research Agency(ANR) as part of the program Investissements d’Avenir (ANR-10-LABX-0083).
{\small\bibliographystyle{ieeenat_fullname}\bibliography{main}}

@String(CVPR= {IEEE Conf. Comput. Vis. Pattern Recog.})

@String(ECCV= {Eur. Conf. Comput. Vis.})

@String(CVPR  = {CVPR})

@String(ECCV  = {ECCV})

@misc{huang2025vipevideoposeengine,
      title={ViPE: Video Pose Engine for 3D Geometric Perception}, 
      author={Jiahui Huang and Qunjie Zhou and Hesam Rabeti and Aleksandr Korovko and Huan Ling and Xuanchi Ren and Tianchang Shen and Jun Gao and Dmitry Slepichev and Chen-Hsuan Lin and Jiawei Ren and Kevin Xie and Joydeep Biswas and Laura Leal-Taixe and Sanja Fidler},
      year={2025},
      eprint={2508.10934},
      archivePrefix={arXiv},
      primaryClass={cs.CV},
      url={https://arxiv.org/abs/2508.10934}, 
}

@inproceedings{NEURIPS2025_336572db,
 author = {Wang, Ruicheng and Xu, Sicheng and Dong, Yue and Deng, Yu and Xiang, Jianfeng and Lv, Zelong and Sun, Guangzhong and Tong, Xin and Yang, Jiaolong},
 booktitle = {Advances in Neural Information Processing Systems},
 doi = {10.52202/085713-1207},
 editor = {D. Belgrave and C. Zhang and H. Lin and R. Pascanu and P. Koniusz and M. Ghassemi and N. Chen},
 pages = {35928--35959},
 publisher = {Curran Associates, Inc.},
 title = {MoGe-2: Accurate Monocular Geometry with Metric Scale and Sharp Details},
 url = {https://proceedings.neurips.cc/paper_files/paper/2025/file/336572db3e99930814d6b328d4220cb6-Paper-Conference.pdf},
 volume = {38, Main Conference},
 year = {2025}
}

@misc{xu2026handflowfullygenerative4d,
      title={HandFlow: Fully Generative 4D Hand Recovery with Flow Matching}, 
      author={Mingxi Xu and Bowen Duan and Yi Gu and Zhengyang Shen and Renjing Xu and Yutao Yue},
      year={2026},
      eprint={2607.11221},
      archivePrefix={arXiv},
      primaryClass={cs.CV},
      url={https://arxiv.org/abs/2607.11221}, 
}

@misc{zhu2026mintunifiedmodelworldspace,
      title={MINT: A Unified Model for World-Space Camera and Hand Motion Estimation from Scalable Egocentric Pipeline Supervision}, 
      author={Zijie Zhu and Weiren Cai and Yizhou Wang and Zhenjie Yang and Yide Liu and Jiahao Chen and Guanqi He},
      year={2026},
      eprint={2609.04958},
      archivePrefix={arXiv},
      primaryClass={cs.CV},
      url={https://arxiv.org/abs/2609.04958}, 
}

@misc{huang2026solarwmopendatascalable,
      title={SolarWM: Open Data and Scalable Training for Long-Horizon Video World Models}, 
      author={Junchao Huang and Guian Fang and Shengju Qian and Xianghao Kong and Zhuoran Zhao and Wei Huang and Yihua Du and Zixin Zhang and Justin Cui and Yuchao Gu and Yukang Chen and Xinting Hu and Tianyu He and Shaoshuai Shi and Zhuotao Tian and Xin Wang and Mike Zheng Shou and Li Jiang},
      year={2026},
      eprint={2609.02886},
      archivePrefix={arXiv},
      primaryClass={cs.CV},
      url={https://arxiv.org/abs/2609.02886}, 
}

@article{li2026cameras,
  title={Cameras as relative positional encoding},
  author={Li, Ruilong and Yi, Brent and Liu, Junchen and Gao, Hang and Ma, Yi and Kanazawa, Angjoo},
  journal={Advances in Neural Information Processing Systems},
  volume={38},
  pages={15984--16009},
  year={2026}
}

@inproceedings{kim2026dexterous,
  title={Dexterous world models},
  author={Kim, Byungjun and Kim, Taeksoo and Lee, Junyoung and Joo, Hanbyul},
  booktitle={Proceedings of the IEEE/CVF Conference on Computer Vision and Pattern Recognition},
  pages={29663--29673},
  year={2026}
}

@misc{xie2026generatedrealityhumancentricworld,
      title={Generated Reality: Human-centric World Simulation using Interactive Video Generation with Hand and Camera Control}, 
      author={Linxi Xie and Lisong C. Sun and Ashley Neall and Tong Wu and Shengqu Cai and Gordon Wetzstein},
      year={2026},
      eprint={2602.18422},
      archivePrefix={arXiv},
      primaryClass={cs.CV},
      url={https://arxiv.org/abs/2602.18422}, 
}

@misc{wang2026hand2worldautoregressiveegocentricinteraction,
      title={Hand2World: Autoregressive Egocentric Interaction Generation via Free-Space Hand Gestures}, 
      author={Yuxi Wang and Wenqi Ouyang and Tianyi Wei and Yi Dong and Zhiqi Shen and Xingang Pan},
      year={2026},
      eprint={2602.09600},
      archivePrefix={arXiv},
      primaryClass={cs.CV},
      url={https://arxiv.org/abs/2602.09600}, 
}

@article{zhu2026causal,
  title={Causal forcing: Autoregressive diffusion distillation done right for high-quality real-time interactive video generation},
  author={Zhu, Hongzhou and Zhao, Min and He, Guande and Su, Hang and Li, Chongxuan and Zhu, Jun},
  journal={arXiv preprint arXiv:2602.02214},
  year={2026}
}

@misc{zhao2026causalforcingscalablefewstep,
      title={Causal Forcing++: Scalable Few-Step Autoregressive Diffusion Distillation for Real-Time Interactive Video Generation}, 
      author={Min Zhao and Hongzhou Zhu and Kaiwen Zheng and Zihan Zhou and Bokai Yan and Xinyuan Li and Xiao Yang and Chongxuan Li and Jun Zhu},
      year={2026},
      eprint={2605.15141},
      archivePrefix={arXiv},
      primaryClass={cs.CV},
      url={https://arxiv.org/abs/2605.15141}, 
}

@article{zhang2026echowm,
  title={EchoWM: Open and enterable omnimodal world models},
  author={Zhang, Songchun and Li, Yaowei and Zhuang, Junhao and Jin, Weiyang and Wang, Haoyu and Lu, Xin and Sun, Yilang and Zhang, Shiyi and Li, Haoran and Ma, Xiaoxiao and others},
  journal={arXiv preprint arXiv:2608.23189},
  year={2026}
}

@InProceedings{Yin_2024_CVPR,
    author    = {Yin, Tianwei and Gharbi, Micha\"el and Zhang, Richard and Shechtman, Eli and Durand, Fr\'edo and Freeman, William T. and Park, Taesung},
    title     = {One-step Diffusion with Distribution Matching Distillation},
    booktitle = {Proceedings of the IEEE/CVF Conference on Computer Vision and Pattern Recognition (CVPR)},
    month     = {June},
    year      = {2024},
    pages     = {6613-6623}
}

@article{huang2026self,
  title={Self forcing: Bridging the train-test gap in autoregressive video diffusion},
  author={Huang, Xun and Li, Zhengqi and He, Guande and Zhou, Mingyuan and Shechtman, Eli},
  journal={Advances in Neural Information Processing Systems},
  volume={38},
  pages={167283--167308},
  year={2026}
}

@InProceedings{Banerjee_2025_CVPR,
    author    = {Banerjee, Prithviraj and Shkodrani, Sindi and Moulon, Pierre and Hampali, Shreyas and Han, Shangchen and Zhang, Fan and Zhang, Linguang and Fountain, Jade and Miller, Edward and Basol, Selen and Newcombe, Richard and Wang, Robert and Engel, Jakob Julian and Hodan, Tomas},
    title     = {HOT3D: Hand and Object Tracking in 3D from Egocentric Multi-View Videos},
    booktitle = {Proceedings of the IEEE/CVF Conference on Computer Vision and Pattern Recognition (CVPR)},
    month     = {June},
    year      = {2025},
    pages     = {7061-7071}
}

@InProceedings{Liu_2022_CVPR,
    author    = {Liu, Yunze and Liu, Yun and Jiang, Che and Lyu, Kangbo and Wan, Weikang and Shen, Hao and Liang, Boqiang and Fu, Zhoujie and Wang, He and Yi, Li},
    title     = {HOI4D: A 4D Egocentric Dataset for Category-Level Human-Object Interaction},
    booktitle = {Proceedings of the IEEE/CVF Conference on Computer Vision and Pattern Recognition (CVPR)},
    month     = {June},
    year      = {2022},
    pages     = {21013-21022}
}

@InProceedings{Yang_2025_CVPR,
    author    = {Yang, Jingkang and Liu, Shuai and Guo, Hongming and Dong, Yuhao and Zhang, Xiamengwei and Zhang, Sicheng and Wang, Pengyun and Zhou, Zitang and Xie, Binzhu and Wang, Ziyue and Ouyang, Bei and Lin, Zhengyu and Cominelli, Marco and Cai, Zhongang and Li, Bo and Zhang, Yuanhan and Zhang, Peiyuan and Hong, Fangzhou and Widmer, Joerg and Gringoli, Francesco and Yang, Lei and Liu, Ziwei},
    title     = {EgoLife: Towards Egocentric Life Assistant},
    booktitle = {Proceedings of the IEEE/CVF Conference on Computer Vision and Pattern Recognition (CVPR)},
    month     = {June},
    year      = {2025},
    pages     = {28885-28900}
}

@misc{lightwheel2026egosuite,
  author       = {{Lightwheel}},
  title        = {{EgoSuite-Open100K: A Fully Annotated Open Egocentric Human Dataset}},
  year         = {2026},
  month        = aug,
  howpublished = {\url{https://huggingface.co/collections/LightwheelAI/egosuite-open100k}},
  note         = {Dataset}
}

@misc{hacohen2026ltx2efficientjointaudiovisual,
      title={LTX-2: Efficient Joint Audio-Visual Foundation Model}, 
      author={Yoav HaCohen and Benny Brazowski and Nisan Chiprut and Yaki Bitterman and Andrew Kvochko and Avishai Berkowitz and Daniel Shalem and Daphna Lifschitz and Dudu Moshe and Eitan Porat and Eitan Richardson and Guy Shiran and Itay Chachy and Jonathan Chetboun and Michael Finkelson and Michael Kupchick and Nir Zabari and Nitzan Guetta and Noa Kotler and Ofir Bibi and Ori Gordon and Poriya Panet and Roi Benita and Shahar Armon and Victor Kulikov and Yaron Inger and Yonatan Shiftan and Zeev Melumian and Zeev Farbman},
      year={2026},
      eprint={2601.03233},
      archivePrefix={arXiv},
      primaryClass={cs.CV},
      url={https://arxiv.org/abs/2601.03233}, 
}

@inproceedings{hu2022lora,
  title = {{LoRA}: Low-Rank Adaptation of Large Language Models},
  author = {Edward J. Hu and Yelong Shen and Phillip Wallis
            and Zeyuan Allen-Zhu and Yuanzhi Li and Shean Wang
            and Lu Wang and Weizhu Chen},
  booktitle = {International Conference on Learning Representations},
  year = {2022},
  url = {https://openreview.net/forum?id=nZeVKeeFYf9}
}

@article{oquab2024dinov2,
  title = {{DINOv2}: Learning Robust Visual Features without Supervision},
  author = {Oquab, Maxime and Darcet, Timoth{\'e}e
    and Moutakanni, Th{\'e}o and Vo, Huy V.
    and Szafraniec, Marc and Khalidov, Vasil
    and Fernandez, Pierre and Haziza, Daniel
    and Massa, Francisco and El-Nouby, Alaaeldin
    and Assran, Mido and Ballas, Nicolas
    and Galuba, Wojciech and Howes, Russell
    and Huang, Po-Yao and Li, Shang-Wen
    and Misra, Ishan and Rabbat, Michael
    and Sharma, Vasu and Synnaeve, Gabriel
    and Xu, Hu and J{\'e}gou, Herv{\'e}
    and Mairal, Julien and Labatut, Patrick
    and Joulin, Armand and Bojanowski, Piotr},
  journal = {Transactions on Machine Learning Research},
  year = {2024},
  url = {https://openreview.net/forum?id=a68SUt6zFt}
}

@inproceedings{zhang2018lpips,
  title = {The Unreasonable Effectiveness of Deep Features
           as a Perceptual Metric},
  author = {Zhang, Richard and Isola, Phillip
            and Efros, Alexei A. and Shechtman, Eli
            and Wang, Oliver},
  booktitle = {Proceedings of the IEEE Conference on
               Computer Vision and Pattern Recognition (CVPR)},
  year = {2018},
  pages = {586--595},
  url = {https://openaccess.thecvf.com/content_cvpr_2018/html/Zhang_The_Unreasonable_Effectiveness_CVPR_2018_paper.html}
}

@article{wang2004ssim,
  title = {Image Quality Assessment:
           From Error Visibility to Structural Similarity},
  author = {Wang, Zhou and Bovik, Alan C.
            and Sheikh, Hamid R. and Simoncelli, Eero P.},
  journal = {IEEE Transactions on Image Processing},
  year = {2004},
  volume = {13},
  number = {4},
  pages = {600--612},
  doi = {10.1109/TIP.2003.819861}
}

@inproceedings{binkowski2018kid,
  title = {Demystifying {MMD GANs}},
  author = {Bi{\'n}kowski, Miko{\l}aj
            and Sutherland, Danica J.
            and Arbel, Michael and Gretton, Arthur},
  booktitle = {International Conference on Learning Representations},
  year = {2018},
  url = {https://openreview.net/forum?id=r1lUOzWCW}
}

@inproceedings{zhang2026controllable,
  title = {Controllable Egocentric Video Generation
           via Occlusion-Aware Sparse {3D} Hand Joints},
  author = {Zhang, Chenyangguang and Ye, Botao
            and Chen, Boqi and Delitzas, Alexandros
            and Wang, Fangjinhua and Pollefeys, Marc
            and Wang, Xi},
  booktitle = {European Conference on Computer Vision (ECCV)},
  year = {2026}
}

@inproceedings{carion2026sam,
  title={Sam 3: Segment anything with concepts},
  author={Carion, Nicolas and Gustafson, Laura and Hu, Yuan-Ting and Debnath, Shoubhik and Hu, Ronghang and Suris Coll-Vinent, Didac and Ryali, Chaitanya and Alwala, Kalyan Vasudev and Khedr, Haitham and Huang, Andrew and others},
  booktitle={International conference on learning representations},
  volume={2026},
  pages={138846--138923},
  year={2026}
}

@misc{gao2026infiniteworldsversatileinteractions,
      title={Infinite Worlds with Versatile Interactions}, 
      author={Zelin Gao and Qiuyu Wang and Jiapeng Zhu and Jingye Chen and Zichen Liu and Qingyan Bai and Jiahao Wang and Yufeng Yuan and Hanlin Wang and Yichong Lu and Ka Leong Cheng and Haojie Zhang and Jian Gao and Tianrui Feng and Yuzheng Liu and Yao Yao and Yinghao Xu and Xing Zhu and Yujun Shen and Hao Ouyang},
      year={2026},
      eprint={2607.07534},
      archivePrefix={arXiv},
      primaryClass={cs.CV},
      url={https://arxiv.org/abs/2607.07534}, 
}

@inproceedings{zhang2026unified,
  title={Unified camera positional encoding for controlled video generation},
  author={Zhang, Cheng and Li, Boying and Wei, Meng and Cao, Yan-Pei and Gambardella, Camilo and Phung, Dinh and Cai, Jianfei},
  booktitle={Proceedings of the IEEE/CVF Conference on Computer Vision and Pattern Recognition},
  pages={38027--38037},
  year={2026}
}

@article{kingma2013auto,
  title={Auto-encoding variational bayes},
  author={Kingma, Diederik P and Welling, Max},
  journal={arXiv preprint arXiv:1312.6114},
  year={2013}
}

@article{zhao2026minwm,
  title={minwm: A full-stack open-source framework for real-time interactive video world models},
  author={Zhao, Min and Zhu, Hongzhou and Yan, Bokai and Zhou, Zihan and Chen, Yimin and Wang, Honglie and Sun, Wenqiang and Fang, Zhengwei and Xun, Zizheng and Li, Zihao and others},
  journal={arXiv preprint arXiv:2605.30263},
  year={2026}
}

@misc{he2026matrixgame20opensourcerealtime,
      title={Matrix-game 2.0: An open-source, real-time, and streaming interactive world model}, 
      author={Xianglong He and Chunli Peng and Zexiang Liu and Boyang Wang and Yifan Zhang and Qi Cui and Fei Kang and Biao Jiang and Mengyin An and Yangyang Ren and Baixin Xu and Hao-Xiang Guo and Kaixiong Gong and Size Wu and Wei Li and Xuchen Song and Yang Liu and Yangguang Li and Yahui Zhou},
      year={2026},
      eprint={2508.13009},
      archivePrefix={arXiv},
      primaryClass={cs.CV},
      url={https://arxiv.org/abs/2508.13009}, 
}

@misc{zhang2025matrixgameinteractiveworldfoundation,
      title={Matrix-Game: Interactive World Foundation Model}, 
      author={Yifan Zhang and Chunli Peng and Boyang Wang and Puyi Wang and Qingcheng Zhu and Fei Kang and Biao Jiang and Zedong Gao and Eric Li and Yang Liu and Yahui Zhou},
      year={2025},
      eprint={2506.18701},
      archivePrefix={arXiv},
      primaryClass={cs.CV},
      url={https://arxiv.org/abs/2506.18701}, 
}

@article{team2026dreamx,
  title={DreamX-World 1.0: A General-Purpose Interactive World Model},
  author={Team, DreamX and Bai, Yancheng and Chen, Rui and Chu, Xiangxiang and Dang, Rujing and Dou, Hao and Gao, Bingjie and Gu, Qiwen and Hong, Siyu and Lei, Jiachen and others},
  journal={arXiv preprint arXiv:2606.16993},
  year={2026}
}

@article{team2026alayaworld,
  title={AlayaWorld: Long-Horizon and Playable Video World Generation},
  author={Team, AlayaWorld and Zhang, Kaipeng and Li, Chuanhao and Zhan, Yifan and Ge, Yongtao and Yin, Yuanyang and Tan, Jiaming and He, Kang and Fan, Liaoyuan and Liu, Ruicong and others},
  journal={arXiv preprint arXiv:2607.06291},
  year={2026}
}

@article{sun2025worldplay,
  title={Worldplay: Towards long-term geometric consistency for real-time interactive world modeling},
  author={Sun, Wenqiang and Zhang, Haiyu and Wang, Haoyuan and Wu, Junta and Wang, Zehan and Wang, Zhenwei and Wang, Yunhong and Zhang, Jun and Wang, Tengfei and Guo, Chunchao},
  journal={arXiv preprint arXiv:2512.14614},
  year={2025}
}

@article{gu2026worldcraft,
  title={WorldCraft: From Camera Navigation to Object Manipulation in Interactive Video World Models},
  author={Gu, Bohai and Wu, Taiyi and Yuan, Yueyang and Liu, Jian and Lu, Xiaocheng and Du, Dazhao and Zhang, Jie and Lai, Jinxiang and Yang, Shuai and Zhao, Xiaotong and others},
  journal={arXiv preprint arXiv:2605.25077},
  year={2026}
}

@inproceedings{bruce2024genie,
  title={Genie: Generative interactive environments},
  author={Bruce, Jake and Dennis, Michael D and Edwards, Ashley and Parker-Holder, Jack and Shi, Yuge and Hughes, Edward and Lai, Matthew and Mavalankar, Aditi and Steigerwald, Richie and Apps, Chris and others},
  booktitle={Forty-first international conference on machine learning},
  year={2024}
}

@inproceedings{ICLR2025_b71ecea2,
 author = {Valevski, Dani and Leviathan, Yaniv and Arar, Moab and Fruchter, Shlomi},
 booktitle = {International Conference on Learning Representations},
 editor = {Y. Yue and A. Garg and N. Peng and F. Sha and R. Yu},
 pages = {73754--73776},
 title = {Diffusion Models Are Real-Time Game Engines},
 url = {https://proceedings.iclr.cc/paper_files/paper/2025/file/b71ecea210f7159f31e46631fe5c838f-Paper-Conference.pdf},
 volume = {2025},
 year = {2025}
}

@misc{rui2026biwmadvancingopensourceinteractive,
      title={BiWM: Advancing Open-Source Interactive Video World Models with Bidirectional Autoregression}, 
      author={Shaohao Rui and Xiaofeng Mao and Zhanyu Zhang and Peijia Lin and Yansong Zhu and Yibo Zhang and Haibin Wan and Zhangrui Zhao and Weijie Ma},
      year={2026},
      eprint={2606.10135},
      archivePrefix={arXiv},
      primaryClass={cs.CV},
      url={https://arxiv.org/abs/2606.10135}, 
}

@article{guo2025mineworld,
  title={Mineworld: a real-time and open-source interactive world model on minecraft},
  author={Guo, Junliang and Ye, Yang and He, Tianyu and Wu, Haoyu and Jiang, Yushu and Pearce, Tim and Bian, Jiang},
  journal={arXiv preprint arXiv:2504.08388},
  year={2025}
}

@inproceedings{che2025gamegen,
  title={Gamegen-x: Interactive open-world game video generation},
  author={Che, Haoxuan and He, Xuanhua and Liu, Quande and Jin, Cheng and Chen, Hao},
  booktitle={International Conference on Learning Representations},
  volume={2025},
  pages={37546--37593},
  year={2025}
}

@article{tang2025hunyuan,
  title={Hunyuan-gamecraft-2: Instruction-following interactive game world model},
  author={Tang, Junshu and Liu, Jiacheng and Li, Jiaqi and Wu, Longhuang and Yang, Haoyu and Zhao, Penghao and Gong, Siruis and Yuan, Xiang and Shao, Shuai and Zhang, Linfeng and others},
  journal={arXiv preprint arXiv:2511.23429},
  year={2025}
}

@article{yang2025longlive,
  title={Longlive: Real-time interactive long video generation},
  author={Yang, Shuai and Huang, Wei and Chu, Ruihang and Xiao, Yicheng and Zhao, Yuyang and Wang, Xianbang and Li, Muyang and Xie, Enze and Chen, Yingcong and Lu, Yao and others},
  journal={arXiv preprint arXiv:2509.22622},
  year={2025}
}

@article{chen2026h3,
  title={H3-world: Turning language understanding into world control},
  author={Chen, Danze and Wang, Zeqing and Lin, Ziyue and Yang, Xingyi and Jin, Yeying},
  journal={arXiv preprint arXiv:2609.01560},
  year={2026}
}

@inproceedings{shin2026motionstream,
  title={Motionstream: Real-time video generation with interactive motion controls},
  author={Shin, Joonghyuk and Li, Zhengqi and Zhang, Richard and Zhu, Jun-Yan and Park, Jaesik and Shechtman, Eli and Huang, Xun},
  booktitle={International Conference on Learning Representations},
  volume={2026},
  pages={7381--7406},
  year={2026}
}

@article{wang2026streaming,
  title={Streaming Video Generation with Streaming Force Control},
  author={Wang, Hanhui and Xie, Yiming and Feng, Haiwen and Lv, Zhaoyang and Wang, Shenlong and Jiang, Huaizu},
  journal={arXiv preprint arXiv:2606.07508},
  year={2026}
}

@article{gillman2026force,
  title={Force prompting: Video generation models can learn and generalize physics-based control signals},
  author={Gillman, Nate and Herrmann, Charles and Freeman, Michael and Aggarwal, Daksh and Luo, Evan and Sun, Deqing and Sun, Chen},
  journal={Advances in Neural Information Processing Systems},
  volume={38},
  pages={103174--103201},
  year={2026}
}

@misc{minimax2026h3,
  author       = {{MiniMax}},
  title        = {{MiniMax H3}: An Open Model Breaking the
                  Boundaries Between Tasks and Modalities},
  year         = {2026},
  month        = jul,
  howpublished = {\url{https://www.minimax.io/blog/minimax-h3}}
}

@inproceedings{akkerman2025interdyn,
  title={Interdyn: Controllable interactive dynamics with video diffusion models},
  author={Akkerman, Rick and Feng, Haiwen and Black, Michael J and Tzionas, Dimitrios and Abrevaya, Victoria Fern{\'a}ndez},
  booktitle={2025 IEEE/CVF Conference on Computer Vision and Pattern Recognition (CVPR)},
  pages={12467--12479},
  year={2025},
  organization={IEEE}
}

@article{chen2026handsonworld,
  title={HandsOnWorld: Unconstrained Egocentric Video Generation with Camera-Disentangled Hand Control},
  author={Chen, Yushuo and Shi, Xiaoyu and Wu, Xiaoshi and Wang, Xintao and Wan, Pengfei and Liu, Yebin},
  journal={arXiv preprint arXiv:2607.02075},
  year={2026}
}

@misc{yan2025openworldhandobjectinteractionvideo,
      title={Open-world Hand-Object Interaction Video Generation Based on Structure and Contact-aware Representation}, 
      author={Haodong Yan and Hang Yu and Zhide Zhong and Weilin Yuan and Xin Gong and Zehang Luo and Chengxi Heyu and Junfeng Li and Wenxuan Song and Shunbo Zhou and Haoang Li},
      year={2025},
      eprint={2512.01677},
      archivePrefix={arXiv},
      primaryClass={cs.CV},
      url={https://arxiv.org/abs/2512.01677}, 
}

@misc{irawan2026ego2actevaluatinggoaldirectedmanipulation,
      title={Ego2Act: Evaluating Goal-Directed Manipulation in Egocentric Video Generation}, 
      author={Patrick Amadeus Irawan and Iskandar Muda Rizky Parlambang and Rava Maulana and Qinrong Cui and Erland Hilman Fuadi and Zayd M. K. Zuhri and Nanda Ryaas Absar and Ahmed Elshabrawy and Wilfried Ariel Mulyawan and Shoubin Yu and Yue Zhang and Mohit Bansal and Alham Fikri Aji},
      year={2026},
      eprint={2610.01092},
      archivePrefix={arXiv},
      primaryClass={cs.CV},
      url={https://arxiv.org/abs/2610.01092}, 
}

@misc{chen2026codeworldmodelcoding,
      title={Code World Model: Coding Agent as World Brain}, 
      author={Yiwen Chen and Guosheng Lin and Chi Zhang},
      year={2026},
      eprint={2608.25927},
      archivePrefix={arXiv},
      primaryClass={cs.CV},
      url={https://arxiv.org/abs/2608.25927}, 
}

@misc{openai2026gpt6,
  title        = {GPT-6 Astra: A New Generation of Intelligence},
  author       = {{OpenAI}},
  year         = {2026},
  howpublished = {\url{https://openai.com/index/gpt-6-astra/}},
  note         = {Accessed: 2026-11-03}
}

@misc{openai2026codex,
  title        = {Code Generation with OpenAI Models and Codex},
  author       = {{OpenAI}},
  year         = {2026},
  howpublished = {\url{https://developers.openai.com/api/docs/guides/code-generation}},
  note         = {Accessed: 2026-11-03}
}
\end{document}